\documentclass{article}

\PassOptionsToPackage{numbers}{natbib}

\usepackage[preprint]{neurips_2023}

\usepackage[T1]{fontenc}    
\usepackage{hyperref}       
\usepackage{nicefrac}       
\usepackage{microtype}      
\usepackage{xcolor}         

\usepackage{nicefrac}       
\usepackage{microtype}      

\usepackage{times}
\usepackage{soul}
\usepackage{url}
\usepackage[utf8]{inputenc}
\usepackage[small]{caption}
\usepackage{graphicx}
\usepackage{amsmath}
\usepackage{amsthm}
\usepackage{booktabs}
\usepackage{algorithm}
\usepackage{algorithmic}
\usepackage{comment}
\usepackage{amsfonts}
\usepackage{subcaption}
\usepackage{amsmath}
\usepackage{amssymb}
\usepackage{bigstrut}

\usepackage{colortbl}
\usepackage{makecell}
\usepackage{dsfont}
\usepackage{alphabeta}

\usepackage{multirow}

\usepackage{mathtools}

\usepackage{xr}
\makeatletter
\newcommand*{\addFileDependency}[1]{
  \typeout{(#1)}
  \@addtofilelist{#1}
  \IfFileExists{#1}{}{\typeout{No file #1.}}
}
\makeatother

\newcommand*{\myexternaldocument}[2][]{%
    \externaldocument[#1]{#2}%
    \addFileDependency{#2.tex}%
    \addFileDependency{#2.aux}%
}

\myexternaldocument[A.]{neurips_2023_appendix}

\usepackage{cleveref}
\crefname{figure}{Figure A.}{Figures A.}
\crefname{table}{Table A.}{Tables A.}
\crefname{section}{Section A.}{Sections A.}

\title{Lucy-SKG: Learning to Play Rocket League Efficiently Using Deep Reinforcement Learning}

\author{%
  Vasileios Moschopoulos\\
  School of Informatics\\
  Aristotle University of Thessaloniki\\
  \texttt{moschopoulos.v@unic.ac.cy} \\
  \And
  Pantelis Kyriakidis\\
  School of Informatics\\
  Aristotle University of Thessaloniki\\
  \texttt{pantelisk@iti.gr} \\
  \And
  Aristotelis Lazaridis\thanks{Corresponding author.}\\
  School of Informatics\\
  Aristotle University of Thessaloniki\\
  \texttt{arislaza@csd.auth.gr} \\
  \And
  Ioannis Vlahavas \\
  School of Informatics\\
  Aristotle University of Thessaloniki\\
  \texttt{vlahavas@csd.auth.gr} \\
}

\begin{document}

\maketitle

\begin{abstract}
A successful tactic that is followed by the scientific community for advancing AI is to treat games as problems, which has been proven to lead to various breakthroughs. We adapt this strategy in order to study Rocket League, a widely popular but rather under-explored 3D multiplayer video game with a distinct physics engine and complex dynamics that pose a significant challenge in developing efficient and high-performance game-playing agents. In this paper, we present Lucy-SKG, a Reinforcement Learning-based model that learned how to play Rocket League in a sample-efficient manner, outperforming by a notable margin the two highest-ranking bots in this game, namely Necto (2022 bot champion) and its successor Nexto, thus becoming a state-of-the-art agent. Our contributions include: a) the development of a reward analysis and visualization library, b) novel parameterizable reward shape functions that capture the utility of complex reward types via our proposed Kinesthetic Reward Combination (KRC) technique, and c) design of auxiliary neural architectures for training on reward prediction and state representation tasks in an on-policy fashion for enhanced efficiency in learning speed and performance. By performing thorough ablation studies for each component of Lucy-SKG, we showed their independent effectiveness in overall performance. In doing so, we demonstrate the prospects and challenges of using sample-efficient Reinforcement Learning techniques for controlling complex dynamical systems under competitive team-based multiplayer conditions.
\end{abstract}

\section{Introduction}

Modeling existing games as environments where a human or system can interact with and be evaluated in a straightforward manner has resulted in significant progress and accomplishments in the domain of Artificial Intelligence (AI) and games, and consequently, AI springs \citep{mcdermott1985dark,thompson2008artificial, 9148327}.
This strategy proved to be ideal for investigating the Reinforcement Learning (RL) paradigm \citep{sutton2018reinforcement}, a sub-field of AI that incorporates the concept of learning through actions and rewards received from an environment to reach set goals. The combination of the latter with Deep Learning opened new pathways for solving problems that were once thought of as too complex to be solved \citep{lazaridis2020deep}.

We opt to focus on Rocket League (Psyonix, 2015), a fast-paced, physics-based 3D online multiplayer car soccer-like game with unique mechanics, where players exercise spatial control and tactics to win.
The game requires players to master its geometry and physics to outmaneuver opponents and maintain control of the ball, as well as a complex spatial skillset compared to other researched games (e.g. Dota, StarCraft). The game also has a different team strategy and positioning than such games, as players must collaborate to score goals and adapt quickly.
These characteristics and complex underlying dynamics make developing an intelligent game-playing system difficult. This, along with the fact that the game is relatively unexplored, were the primary reasons that attracted our interest.

In this paper, we present Lucy-SKG (\textbf{S}haping \textbf{K}inesthetic Intelli\textbf{G}ence), an agent that learns to play Rocket League efficiently using Deep Reinforcement Learning and outperforms Necto, the Rocket League Bot Championship 2022 winner \citep{necto, RLBotChampionshipFinale}, with significantly reduced training times. We show that our proposed novelties lead to sample-efficient learning and high overall performance in the game, leading to constant winning streaks against Necto, as well as Nexto, the successor of Necto.

Our work is the first thorough investigation in the commercial game of Rocket League, exploring a new pathway for competitive team-based control problems of complex physics systems, non-linear dynamics and action space, offering unique characteristics in comparison to other games. Our contributions and novelties are summarized as follows:

\begin{itemize}
    \item Development of a reward analysis and visualization library for Rocket League
    \item Proposal and use of Kinesthetic Reward Combination (KRC), an alternative to linear reward combinations and useful in measuring utility of complex phenomena
    \item Study of reduced amount of information and the effect of previous action stacking for the state space
    \item Design of novel, parameterized reward functions, applicable to similar field games
    \item Implementation and study of on-policy neural architectures trained on unsupervised auxiliary tasks in a complex 3D simulation environment (Rocket League)
    \item Thorough ablation studies for each of Lucy-SKG component and their performance impact on the agent
    \item A new state-of-the-art bot for Rocket League.
\end{itemize}

In the next sections we explore previous related work (Section \ref{related}), we briefly introduce Rocket League and technical preliminaries crucial for this work (Section \ref{background}), we describe in detail our proposed model (Section \ref{methodology}), present the experiments conducted and their results (Section \ref{experiments}), discuss on our findings (Section \ref{discussion}) and conclude our work in Section \ref{conclusion}, where we sum up our proposed model and its capabilities, as well as give insights regarding future directions.\footnote{Figures, Tables and Sections referenced throughout the paper with the prefix `A.' refer to the Appendix.}

\section{Related Work}
\label{related}

Until recently, the Rocket League bot-developing community was limited to hard-coded bots with no learning capabilities, or supervised learning approaches for predicting match outcomes and player ranks \citep{smithies2021random}. Although traditional RL methods appeared to fail due to the complexity of the environment \citep{berman2021exploration}, \textit{Element} was the first Deep RL-based bot with a satisfying learning mechanism. \textit{Necto} followed afterwards and, by employing more sophisticated techniques, it achieved state-of-the-art performance regarding Rocket League bots, having won the Rocket League Bot Championship 2022. Necto is also a milestone in reward shaping for multiplayer physics games, inspired by \citet{liu2021motor} for its reward function.

Recently, \textit{Nexto}, Necto's successor (also known as \textit{Necto v2}), took the spotlight. Nexto became notorious to the Rocket League community for its public and extensively pre-trained version that was widely used within ranked games for winning against top-rank human players. To date, it is the highest-ranking bot, followed by Necto. Its key differences to Necto are: \textit{a)} an action space with explicit rules that prohibit certain invalid action combinations that remove learning load and substitute it with explicit human knowledge (not to be confused with action masking/illegal actions, which are decided by the environment), \textit{b)} a tweaked version of Necto's reward function (it uses similar reward components to Necto and modified reward weights) and c) more network parameters. We regard Nexto as a fine-tuned version of Necto, rather than a different bot, which is why we use Necto as our main baseline. In our experiments, we evaluate Lucy-SKG against both Necto and Nexto.

A more recent benchmark that was established is GT Sophy \citep{wurman2022outracing}, a Gran Turismo agent. Despite the game being a racing game, Gran Turismo shares only few similarities with Rocket League. Although GT Sophy excels in a regular racing setting, the game itself is not designed for performing complex aerial maneuvers and employing team coordination skills such as the ones required in Rocket League. We argue that handling the ball, performing actions such as intentional shots, passes, saves and proper positioning (on-ground and mid-air) accurately in Rocket League is a much more difficult task to learn. Furthermore, performing in a team-based tactical manner within a restriction-free and open-field environment with rich game mechanics and complex action space presents an additional challenge. Therefore, the overall experience between the two games is vastly different.

In this work, we employ novel and reward shaping techniques, generalizable to other games, such as sports games (FIFA, NBA), being assisted by insights through reward analysis. Moreover, we incorporate on-policy auxiliary task methods---contrary to the off-policy methods by \citet{yarats2021improving} and \citet{jaderberg2016reinforcement}---to further improve sample efficiency and performance. Additionally, we present positive findings from the use of a simplified observation space that does not include all available information from the environment.

\section{Background}
\label{background}


\subsection{Rocket League}
\label{rocketleague}

Rocket League is a 3D multiplayer soccer-like game developed in Unreal Engine \citep{unrealengine}, where each player controls a vehicle, maneuvering quickly within the bounds of a closed stadium (Figure A.1). Matches can be 1-vs-1 and go up to 4-vs-4. The goal is to place a large soccer ball into the opponents' goal by hitting it; the team with the most goals within 5 minutes wins the game. Players can also collect boost points from boost pads found in fixed locations on the field, used for temporary speed boosts and aerial moves. RL bots for Rocket League are usually developed using RLGym \citep{rlgym}, a framework that provides an OpenAI Gym \citep{1606.01540} interface for the game.

\subsection{Reward Shaping}
\label{rewardbg}

A typical RL-formulated problem is defined by a Markov Decision Process (MDP) with a 5-tuple $M = (S, A, p, \gamma, R)$, where $S$ is the state space, $A$ is the action space, $p$ is the environment dynamics function describing the probability $Pr\{S_t=s', R_t=r|S_{t-1}=s, A_{t-1}=a\}$, $\gamma$ is a discount factor and $R$ is the reward function \citep{sutton2018reinforcement}.

A common problem for many MDPs is that $R$ is often very sparse, providing non-zero rewards to the agent for few states only \citep{minsky1961steps, sutton2018reinforcement} and making learning difficult. This phenomenon is also present in Rocket League, since several significant rewarding events take place only a few times during a match, if any at all. For instance, a ``Goal'' event is a highly rewarding case, but is difficult for the agent to determine accurately the chain of events that led to it.

A solution to this, proposed by \citet{ng1999policy}, is to transform the original MDP $M$ into a shaped MDP $M' = (S, A, p, \gamma, R')$, where \(R' = R + F \) and \(F(s, a, s') = \gamma \Phi(s') - \Phi(s)\). In this case, $R'$ is a shaped reward function, $F: S \times A \times S \mapsto \mathbb{R}$ is a bounded, real-valued potential-based reward shaping function, and $\Phi: S \mapsto \mathbb{R}$ is a potential function that measures a given state's quality.

This technique helps agents learn and converge quicker by densifying the reward function using function $F$, hence providing intermittent reward signals that better inform agents about the quality of visiting states, compared to the case with sparse reward signals. When $\Phi$ is continuous, $F$ is continuous as well, introducing a gradient in the reward function that agents can use in order to learn and converge faster.

Due to the game's nature and complex dynamics that require multiple skills, an efficient Rocket League environment setup should support various types of extrinsic rewards that correspond to different in-game events and skills. For this reason, we define the agent's reward function $R$, as well as the potential function $F$, to be weighted linear combinations of various reward function components. The weights for each component were attributed by empirical experimentation: balancing their effects in the reward function after gaining insights from gameplay analyses, visualizing reward functions, as well as considering their value ranges in the arena. Therefore, these functions can be described as:

\[R = w_{R_1} R_1 + w_{R_2} R_2 + \dots + w_{R_n} R_n, \quad
\Phi = w_{\Phi_1} \Phi_1 + w_{\Phi_2} \Phi_2 + \dots + w_{\Phi_m} \Phi_m,\]

where $R_i$ and $\Phi_j$ are reward functions $i$ and $j$ with $w_{R_i}$ and $w_{\Phi_j}$ being their corresponding weights, $n$ the number of reward function components used for the reward function and $m$ the number of reward function components used for the potential function.

Additionally, due to the fact that reward function components used for $\Phi$ measure the quality of a given state, we refer to $\Phi$ as \textit{general utility}, and reward functions $\Phi_i$ as \textit{utilities}. Furthermore, reward functions $\Phi_i$ can be further divided into \textit{state utilities} and \textit{player utilities}, that measure the quality of the state attributed to non-player and player properties, respectively.

\subsection{Auxiliary Tasks}
\label{unreal}
In order to enhance the sample efficiency and performance of our model, we employed auxiliary task methods that aim in learning to predict certain parts of the environment, such as the state and the reward. Auxiliary task methods aim to regularize the main-objective network parameters, i.e. playing Rocket League, with respect to auxiliary goals.

Using similar formulation to \citet{jaderberg2016reinforcement}, we define each auxiliary task as $c$ and the reward function with respect to task $c$ as $R^{c}: \mathcal{S} \times \mathcal{A} \to \mathbb{R}$. The objective (eq. \ref{auxObj}) is the maximization of expected discounted returns, both for the main and auxiliary tasks, using a weight $\lambda_c$ that controls the impact of each auxiliary task on the total loss.

\begin{equation}
    \operatorname*{arg\,max}_\alpha \mathbb{E}_\pi \sum_{k=0}^\infty \gamma^k R_{t+k+1} + \sum_{c\in C} \lambda_c \mathbb{E}_\pi \sum_{k=0}^\infty \gamma^k R^{(c)}_{t+k+1}
    \label{auxObj}
\end{equation}


The loss function for optimizing the agent's networks consists of the main objective loss $\mathcal{L}_{RL}$ and the weighted auxiliary losses, and is defined as \( \mathcal{L}_{\theta} = \mathcal{L}_{RL} + \sum_{c\in C} \lambda_c \mathcal{L}^{(c)} \label{auxLoss}\), where $\theta$ refers to the networks' learnable parameters.

\section{Methodology}
\label{methodology}


\subsection{Rocket League Reward Analysis Library}
\label{rewardlib}

For the purposes of studying and visualizing reward functions within a typical Rocket League arena, we developed the \textit{rlgym-reward-analysis} Python library. The library contains two main modules: \textit{1.} the \textit{visualization module}, which allows users to create contour plots of reward functions for Rocket League, and \textit{2.} the \textit{replay file-to-reward} module, used for extracting reward values from frames in Rocket League replay files generated by the Carball \citep{carball} library. By studying the existing RLGym reward functions, assisted by the visualization module, we detected the following weaknesses:
\begin{itemize}
\item \textbf{Distance rewards cannot be parameterized}. An example of this is `Ball-to-Goal Distance' which produces the same static image when visualized.
\item \textbf{`Align Ball-to-Goal` does not take into account player distance}. When visualized, the function has the shape of a ``beam`` (Figure A.4).
\item \textbf{`Player-to-Ball Velocity` is inherently less potent as a reward}. When the player moves toward the ball it may simply hit it toward the team goal.
\item \textbf{`Touch Ball Acceleration`, a reward exclusive to Necto, may be improved} by computing acceleration toward the opponent goal, i.e. `Touch Ball-to-Goal Acceleration`. This doubles function range by introducing direction and penalizes for hitting the ball toward team goal.
\end{itemize}

This analysis allowed us to gain useful insights and perform improvements over existing reward functions, as well as develop our own novel ones. These are described in detail in Section \ref{Lucy-SKGrewards}.

\subsection{Architecture}
\label{arch}

For the architecture of Lucy-SKG, we used the same as Necto, due to the fact that it is a relatively simple and commonly used cross-attention architecture that is known to work effectively, allowing us to focus on our proposed novelties. Additionally, it would allow us to perform more straightforward comparisons during ablation studies and experiments against Necto.
This architecture uses the Perceiver \citep{jaegle2021perceiver}, cross-attention mechanism and consists of two identical copies for the actor and the critic. Each copy contains two MLP layers for preprocessing the latent and the byte array, followed by 2 Transformer \citep{vaswani2017attention} encoder layers that perform cross-attention. 

Preprocessing layers of the actor network were branched into separate networks, for auxiliary task learning. Each branch was connected to the preprocessing layers via a corresponding multi-head cross-attention layer. The architecture incorporates Proximal Policy Optimization (PPO) \citep{schulman2017proximal} as the learning algorithm, since it has already been proven to be an efficient Deep Reinforcement Learning algorithm for cooperative and multi-agent games \citep{yu2022the}, such as in OpenAI Five \citep{openaifive}. 
Other present design choices are elaborated in Section A.5.

\subsection{Observation space}
\label{obsspace}

Due to the architecture, the observation space needs to be constructed as a triplet of a latent array (\textit{query}) containing information about the current player, a byte array (\textit{key/value}) that represents a sequence of objects to pay attention to, and a \textit{key padding mask} that allows the agent to train on a vectorized environment of match instances with different number of players. This type of architecture can also be used in all types of environments where players get to interact with objects, players and NPCs that share the same type of features.

For Lucy-SKG, we define the state space as a portion of the state space available through RLGym, contrary to other agents of the game (Necto, Element) or other games (GT Sophy); this could potentially be a step forward towards future implementations that use raw pixel input.
This decision was made for the purposes of:
\textit{a)} studying a less rich state space and enhancing it with previous agent actions,
\textit{b)} reducing memory usage with the goal of fitting batch sequences, necessary for the reward prediction auxiliary task, in memory, and
\textit{c)} reducing the computational complexity by reducing the number of key/value objects.

The state space for Lucy-SKG is given in detail in Section A.1.2, but below are the main differences between it and the full state space:

\textbf{Lack of boost pad objects.} Lucy-SKG is not informed of boost pad information which reduces the observation to 8 objects. This way, we are allowed to fit batch sequences in memory that are required for the reward prediction auxiliary task. An additional intuition behind this decision is that boost pads are static objects with fixed positions and the agent should be able to discover them without significant difficulty, focusing on fewer objects at each time step.
\\\textbf{Lack of demolition/boost timers.} 
Compared to Necto, Lucy-SKG uses a demolished flag only instead of demolition/boost timers. This way, we enhance the generalization abilities of the model by avoiding to explicitly specify refresh times.
\\\textbf{Previous-action stacking.} For query features only, the $k$ previous actions of the player are additionally appended in order for the agent to become more aware of their consequences. In our implementations, we set $k=5$.

\subsection{Reward Shaping in Lucy-SKG}
\label{Lucy-SKGrewards}

\subsubsection*{Kinesthetic Reward Combinations}

Traditional reward shaping often takes place in the form of linear reward combinations that comprise a weighted sum over individual reward components, as presented in subsection \ref{rewardbg}. A disadvantage of such combinations is the reward signal may vary, depending on the magnitude of individual components, making it harder for agents to distinguish the effects of their actions and learn optimal policies in complex environments, such as Rocket League.

To combat this, we propose a novel way of combining reward functions, the \textit{Kinesthetic Reward Combination} (KRC), defined as:

\begin{equation}
\resizebox{0.6\textwidth}{!}{
    \({R_{c}} = sgn \left(r \right) * {\sqrt[n]{{\prod\limits_{i = 1}^n {\left| {{R_i}} \right|} }}}, \quad
    sgn \left(r \right) = \begin{cases*}
    $1$ & if $r > 0, \forall r \in R$\\
    $-1$ & otherwise
    \end{cases*}\),
}
\label{equation1}
\end{equation}

where $R$ is the set of the individual reward components. The case $sgn(0)$ is trivial since \({{\prod\limits_{i = 1}^n {\left| {{R_i}} \right|} }}=0\). Intuitively, this type of combination was defined as such so as to take into account $n$ reward signals altogether and scale them using $n$-th root transformation in order to inflate smaller values and stabilize larger ones. Additionally, the $sgn$ function controls the resulting reward's sign, providing positive rewards only when all individual rewards are positive. The KRC could potentially be generalized to support weight attribution by applying powers to absolute component values and taking the corresponding root, allowing certain components to have a greater effect. This case, however, requires further analysis and is left as future work.

The advantage of this combination is that the resulting reward signal is compound, representative of a single, mixed reward function and indicative of high-level state quality information.
In this work, a mixture of linear combinations and KRCs is employed for the reward function, using KRCs as components of a linear one. The reason for this is that using linear combinations only, multiple simple components are maximized independently whereas several complex skills should be learned instead. An example of such a skill is aligning the ball toward the goal while maintaining a close distance to it. On the other hand, constructing a single large reward using a KRC attenuates the effect of various components when too many of them are used.

\begin{figure}[t]
    \centering
    \begin{subfigure}[b]{0.24\textwidth}
        \includegraphics[width=\linewidth]{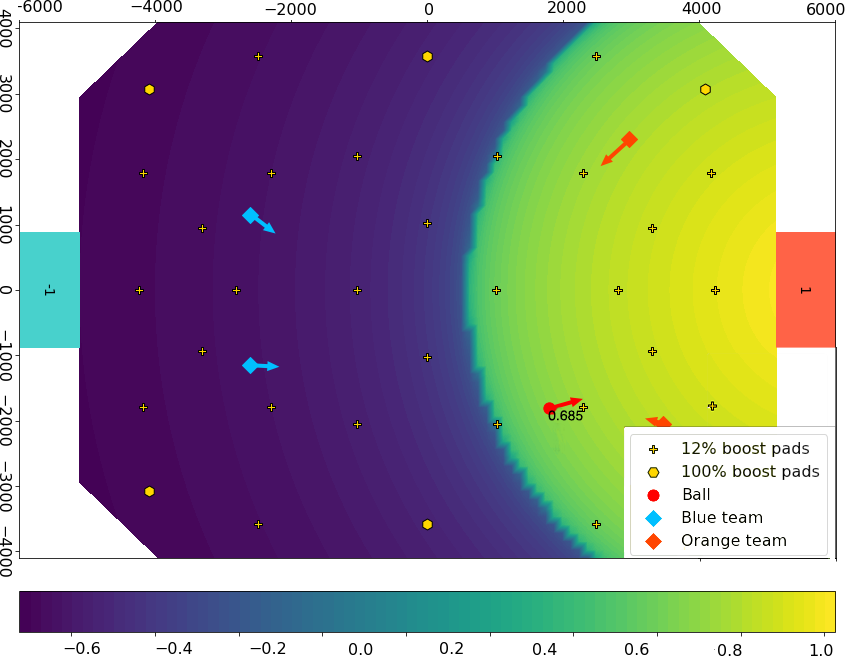}
    \end{subfigure}
    \begin{subfigure}[b]{0.24\textwidth}
        \includegraphics[width=\linewidth]{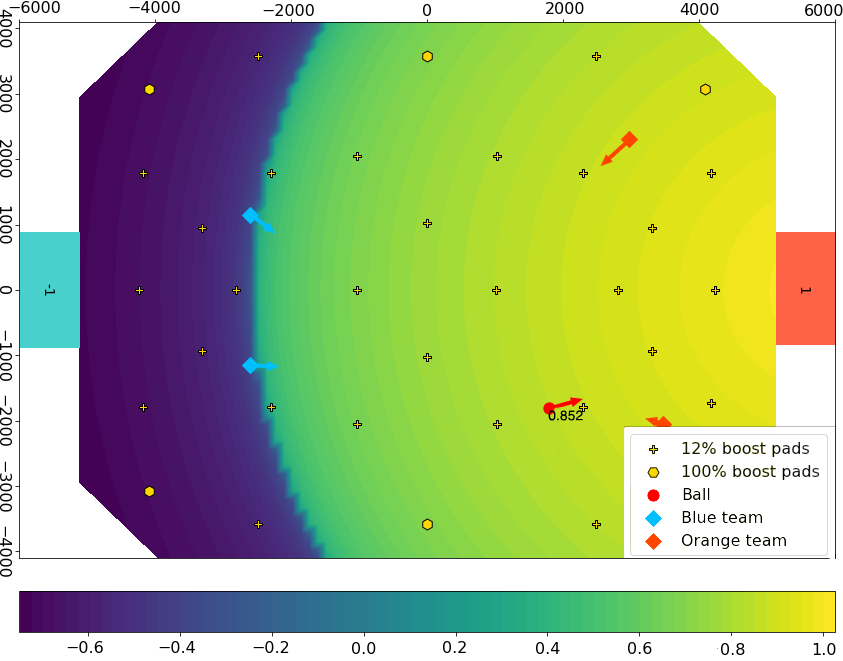}
    \end{subfigure}
    \begin{subfigure}[b]{0.50\textwidth}
        \includegraphics[width=\linewidth]{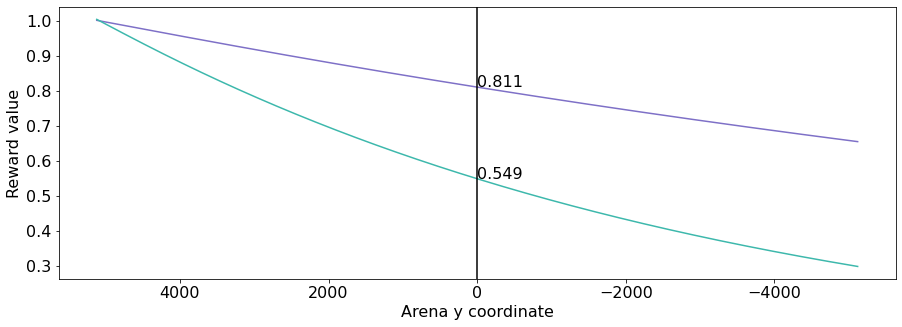}
    \end{subfigure}
    \caption{Left: Effects of dispersion factor when set to 0.7 and 1.2 on the `Signed ball-to-goal distance' reward function. Density set to 5 for visualization purposes. Right: Effects of density when $w_{dis} = 0.7$ (bottom curve) and $w_{dis} = 2$ (top curve) on `Ball-to-Goal distance' reward function. Arena x and z coordinates set to 0.}
    \label{fig:dispersion comparison}
\end{figure}


\subsubsection*{Parameterized distance reward functions}

In order to improve existing distance reward functions, we parameterized them by introducing \textbf{i)} a \textit{dispersion} factor $w_{dis}$ that controls distance reward spread and allows larger reward weights to extend further from (or gather closer toward) the distance reward target and \textbf{ii)} a \textit{density} factor $w_{den}$ that increases and decreases reward values within the function’s range, by controlling concavity. A distance reward function parameterized through dispersion and density, can be described as follows:

\begin{equation}
    \resizebox{0.4\textwidth}{!}{
    \(    R_{dist} = \exp\left(-0.5 * \frac{d(i,j)}{c_d * w_{dis}}\right)^{1 / w_{den}} \),
    }
    \label{eq:parameterized distance fun}
\end{equation}

where $d$ is a distance function (e.g Euclidean distance) between two objects $i$ and $j$, and $c_d$ its normalizing constant.  A visual example of these two parameters' effects can be seen in Figures \ref{fig:dispersion comparison}.

For our implementation, we used both novel and existing reward functions that are either utility functions, as defined in subsection \ref{rewardbg}, or event reward functions, which correspond to in-game events. All utility reward function components used are in $[0, 1]$ or $[-1, 1]$ ranges, ensuring fair reward weight attribution. Rewards are distributed among players using a \textit{Team Spirit} factor $\tau$, a technique similar to the one used in OpenAI Five \citep{berner2019dota}.

The main characteristics of the reward function are: 
\begin{itemize}
    \item \textbf{Use of KRCs.} Through KRCs, we introduce ``Offensive Potential" and ``Distance-weighted Alignment".
    \item \textbf{Distance reward parameterization.} We parameterize distance reward functions by dispersion and density, as defined in eq. \ref{eq:parameterized distance fun}.
    \item \textbf{Careful selection of components.} All the reward components used were carefully selected using our analysis library.
    \item \textbf{Modification of reward functions.} ``Touch Ball Acceleration" was replaced by ``Touch Ball-to-Goal Acceleration".
\end{itemize}

The reward function is described in detail in Section A.1.1.

\subsection{Auxiliary Task Learning}
\label{aux}

Our auxiliary task learning methodology is based on UNREAL \citep{jaderberg2016reinforcement}, an agent that creates rich representations by using auxiliary networks to optimize pseudo-reward functions related to the main objective. Each auxiliary task provides a specific pseudo-reward that guides the agent towards learning a part of the environment. In this work, we employ 2 such tasks: \textit{State Representation} and \textit{Reward Prediction}. Both task networks share the same preprocessing layers of a network or the shared layers of an actor-critic network (Section A.2).

Contrary to the off-policy A3C \citep{mnih2016asynchronous} algorithm of UNREAL, we opted for the on-policy auxiliary training since we chose PPO as our learning algorithm. By having access to replay memory, UNREAL uses n-step Q-learning to define the auxiliary losses for most tasks. On the other hand, we directly plug in the self-supervised losses.

State Representation (SR) is the task of creating accurate environment reconstructions, using Autoencoder neural networks which compress the input to a low-dimensional space and subsequently recreate it \citep{rumelhart1985learning, bank2020autoencoders}.
We use ``smooth L1" \citep{girshick2015fast} as reconstruction loss, an L1 variant with L2-like behavior for small values, which makes it less sensitive to outliers and more balanced against other losses.

Reward Prediction (RP) refers to predicting immediate reward value (like a critic with $\gamma = 0$) given a sequence of states. In the environments originally studied in UNREAL, reward signals are sparse and RP assists the agent in visiting rewarding states by exploring the environment efficiently. In contrast, our reward function is dense (Section \ref{Lucy-SKGrewards}). Nevertheless, the benefit is the conditioning of the agent to an observation sequence---instead of single ones--- with respect to immediate rewards, the fast convergence to these types of rewards, and by extension to other event-based ones.

For observation space $\mathcal{O} \subseteq \mathcal{S}$, let $\mathbf{o_t} \in \mathcal{O}$ be a $d$-dimensional observation sampled at timestep $t$ and $r_{t+1} \in \mathbb{R}$ be the received reward given by the environment reward function $R$ after performing action $\mathbf{a_t} \in A$ in state $\mathbf{s_t}$. For the RP task we create a history of the $l$ previous observations to $\mathbf{o_t}$, i.e. $\mathbf{H_t} = \{\mathbf{o_{t-(l-1)}}, ..., \mathbf{o_t}\} \in \mathcal{H}^{l\times d}$ for the sampled $\mathbf{o_t}$, where $l$ is the chosen sequential length and $\forall \mathbf{o_j} \neq \mathbf{o_t}$ in $\mathbf{H_t}$ that is not available (e.g. in episode starts) we use a zeroed-out observation.

RP aims to do a non-linear mapping from $\mathbf{H_t}$ to an one-hot encoded output $y_t \in \mathcal{Y}^{\{0, 1\}\times 3}$ using a Neural Network function $\phi: \mathcal{H} \to \mathcal{Y}$. Output dimensions indicate the multi-class classification of $r_t$ as positive, negative, or near-zero. Since we have sequences of observations, we opt to use an LSTM-based architecture \citep{hochreiter1997long} with a fairly large sequence length to account for the game's complexity.

\section{Experimental Results}
\label{experiments}

\begin{figure*}[t]
  \centering
    \begin{subfigure}[b]{0.28\textwidth}
        \includegraphics[width=\textwidth]{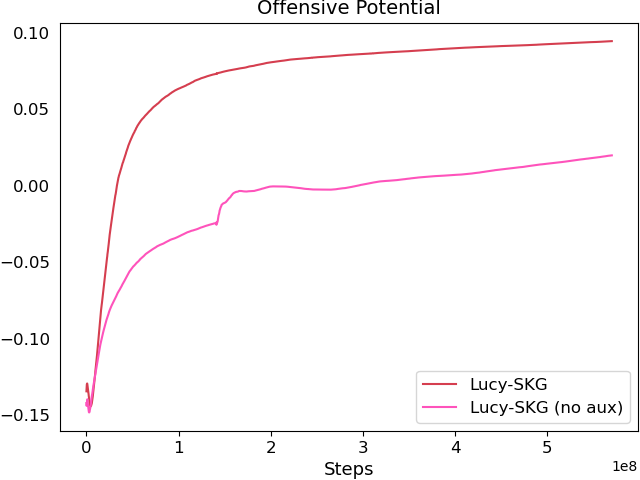}
        \label{fig:offensive potential}
    \end{subfigure}
    \begin{subfigure}[b]{0.28\textwidth}
        \includegraphics[width=\textwidth]
        {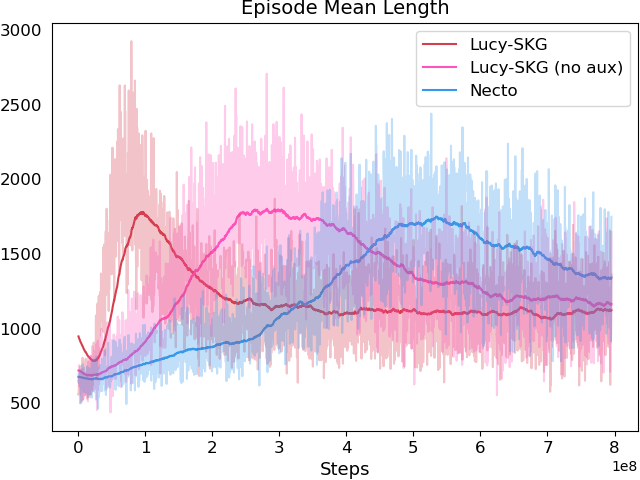}
        \label{fig:episode mean length}
    \end{subfigure}
    \begin{subfigure}[b]{0.28\textwidth}
        \includegraphics[width=\textwidth]
        {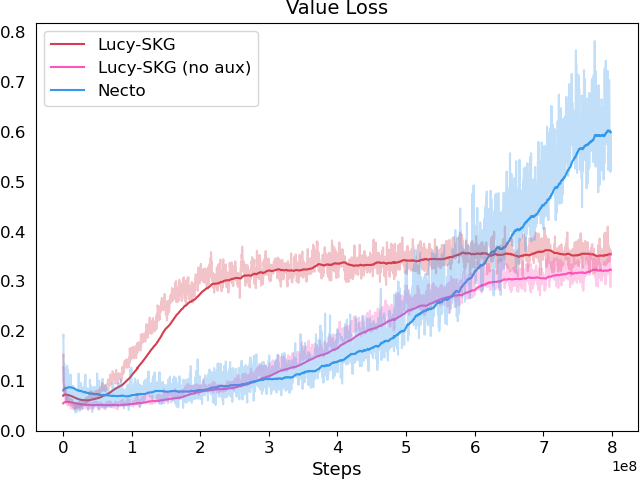}
        \label{fig:value loss}
    \end{subfigure}

  \caption{\centering Performance evaluation metrics for Lucy (with \& without using auxiliary task methods) and Necto.}
  \label{lucy_skg_rewards}

  \centering
    \begin{subfigure}[b]{0.28\textwidth}
        \includegraphics[width=\textwidth]{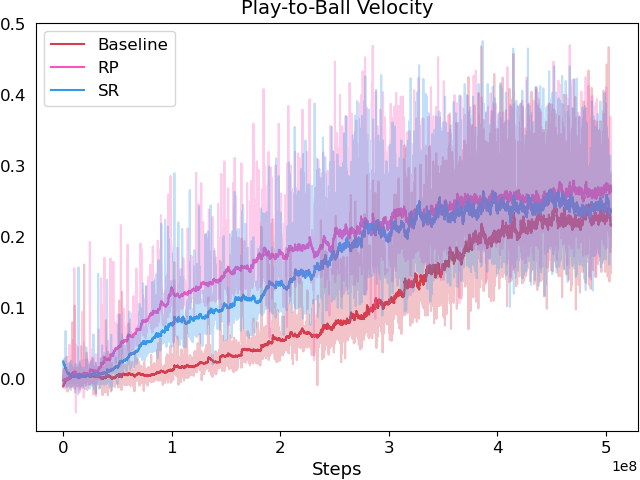}
        \label{fig:aux save}
    \end{subfigure}
    \begin{subfigure}[b]{0.28\textwidth}
        \includegraphics[width=\textwidth]
{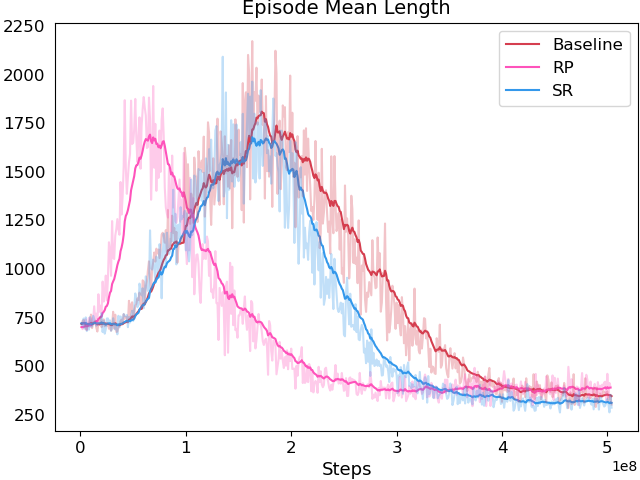}
        \label{fig: aux episode mean length}
    \end{subfigure}
    \begin{subfigure}[b]{0.28\textwidth}
        \includegraphics[width=\textwidth]
        {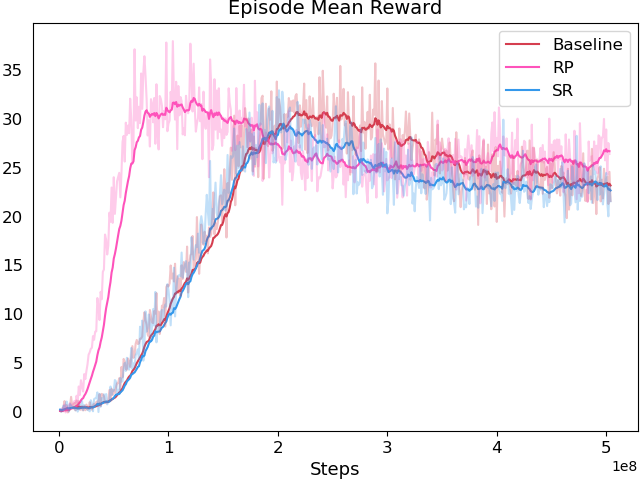}
        \label{fig:aux episode mean reward}
    \end{subfigure}

  \caption{\centering Performance evaluation metrics for auxiliary task methods.}
  \label{aux_rewards}
\end{figure*}

\subsection{Auxiliary Tasks Results}
\label{aux_results}

For the auxiliary task components, we used a simple architecture as a baseline\footnote{https://github.com/Impossibum/rlgym\_quickstart\_tutorial\_bot} (Section A.2) developed by the RLGym community. We compared against the baseline, by plugging to the baseline network each auxiliary network, using the same hyper-parameters for comparison purposes. 

Results indicate (Fig. \ref{aux_rewards}) that the RP task is more sample-efficient (Figure A.6) for all rewards (Table A.1), and performs better in most cases compared to the baseline. More specifically, in the mean episode length it is the first to reach a peak and stabilize faster, meaning that it exploits immediate rewards fully (peak) and learns to score faster (stabilization). Its sample-efficiency is evident in the mean episode reward as well,  hitting a relative plateau at $100$M steps, whereas the baseline hits a plateau around $200$M. Moreover, it outperforms both SR and the baseline in terms of mean episode rewards and `Player-to-Ball Velocity'. For SR (Fig. \ref{aux_rewards}), results do not appear to follow a similar trend, since the agent struggles to outperform the baseline in both Mean Episode Length and Reward. However, SR has an edge in terms of peak value and sample-efficiency in ``Player-to-Ball Velocity" and appears to be superior in the Demolition reward (Figure A.7).

\subsection{Reward Shaping Results}
\label{reward_shaping_results}

To measure the effects of our reward shaping techniques, we studied the progression of the episode mean length and the value loss during training (Fig.~\ref{lucy_skg_rewards}). Lucy-SKG (no aux) displayed an upward trend in episode mean length earlier during training compared to Necto. Episode mean length increases as the agent learns to touch the ball, thus increasing return by extending the episode during initial training stages, and subsequently decreases as it learns that scoring sooner provides the most return.

On the other hand, value loss increased earlier as well with Lucy-SKG (no aux), signifying an increase in action novelty. At about 600M steps, value loss followed an upward trend for Necto, interpreted as its actions being less predictable due to its reward function's nature. This presents certain advantages for Lucy-SKG, meaning a more consistent critic and a more powerful reward function overall.

The benefits are attributed to the use of the `Offensive Potential' and `Distance-weighted Alignment' KRCs, which displayed a notable increase early on. The above findings align with Lucy-SKG, as well, and are greatly enhanced by auxiliary task learning.

\subsection{Lucy-SKG vs. Necto \& Nexto Evaluation Results}
\label{combined_results}

\begin{figure}[tbp]
    \centering
  \begin{minipage}[b]{0.47\textwidth}
    \centering
    \includegraphics[width=1\textwidth]{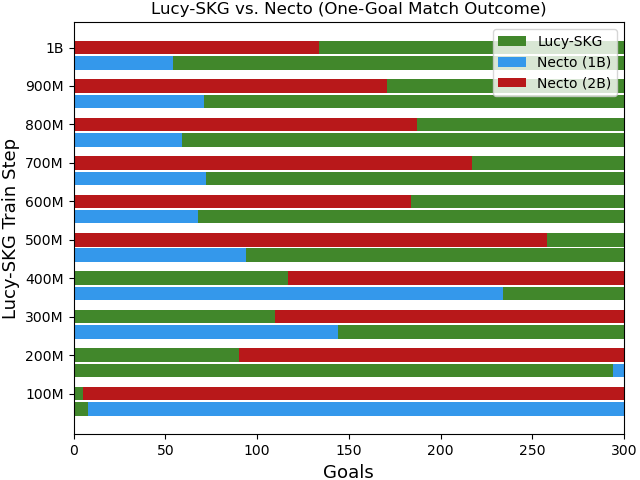}
    \captionof{figure}{Lucy-SKG vs Necto match results.
    }
    \label{evaluation_games}
  \end{minipage}%
  \vspace{5pt}
  \begin{minipage}[b]{0.4\textwidth}
    \centering
    \resizebox{1\textwidth}{!}{%
    \begin{tabular}{|c|c|c|c|}
        \hline
        \textbf{Team 1 (Blue)} & \textbf{Team 2 (Orange)} & \textbf{Outcome} & \textbf{Win \%} \bigstrut\\
        \hline
        Lucy-SKG & Nexto & 300-4 & 100\% 
        \bigstrut\\
        \hline
        \hline
        Lucy-SKG & Necto & 300-54 & 100\% 
        \bigstrut\\
        \hline
        Lucy-Gamma & Necto & 300-200 & 97.88\%
        \bigstrut\\
        \hline
        \hline
        Lucy-Gamma & Lucy-SKG & 73-300 & 0\% \bigstrut\\
        \hline
        Lucy-Beta & Lucy-Gamma & 158-300 & 0.20\% \bigstrut\\
        \hline
        Lucy-Alpha & Lucy-Beta & 4-300 & 0\% \bigstrut\\
        \hline
        \hline
        Necto Reward & Necto & 300-220 & 100\% \bigstrut\\
        \hline
        Necto Reward & Lucy-SKG & 103-300 & 0.24\% \bigstrut\\
        \hline
        Necto Reward & Lucy-Gamma & 283-300 & 2.64\% \bigstrut\\
        \hline
        Necto Reward & Lucy-Beta & 300-129 & 99.77\% \bigstrut\\
        \hline
        \hline
        Necto & Nexto & 300-4 & 100\% \bigstrut\\
        \hline
    \end{tabular}
    }
    \captionof{table}{Evaluation between Lucy-SKG, Necto and Nexto at 1B steps. \textit{Win \%} is with respect to Team 1.
    }
    \label{tab:evaluation results}
  \end{minipage}
\end{figure}

For our direct evaluation against Necto and Nexto we recorded the total score on 300 independent single-goal (i.e. first scorer wins) 2-vs-2 matches and present the results in Table \ref{tab:evaluation results}. We also include the percentage of matches that Team 1 was ahead in total score. All models were trained for 1B steps.

The results are satisfying, with Lucy-SKG winning most matches against Necto (300:54), displaying its enhanced learning capabilities. Additionally, Lucy-SKG showed superior performance against Nexto as well (300:4). This indicates that Nexto lacks learning efficiency, also evidenced by its poor performance against Necto. This can be attributed to it being a poorly tweaked version of Necto, plus it is possible that its larger architecture consequently leads to slower learning. However, we believe 1B training steps displays accurately its lack of sample-efficiency, and can be a representative indication of its overall performance. Moreover, we trained Necto at 2B steps, and evaluated it against Lucy-SKG trained at 100M time step intervals. Results showed that Lucy-SKG managed to win from 200M and 400M time steps afterwards against the 1B and the 2B Necto versions respectively, showing \textasciitilde5x improved learning speed (Figure \ref{evaluation_games}). Detailed numerical results are given in Table A.7.

\subsection{Ablation Study}
\label{ablation study}

In order to study the individual impact of each of Lucy-SKG's proposed components in the overall performance, we conducted evaluation experiments using the following trained versions of our model:
\begin{itemize}
    \item \textbf{Lucy-SKG:} Final model.
    \item \textbf{Lucy-Gamma:} Lucy-SKG without using auxiliary task methods.
    \item \textbf{Lucy-Beta:} Lucy-Gamma without using our reward shaping methodology (replaced it with Necto reward). 
    \item \textbf{Lucy-Alpha:} Lucy-Beta without 5-action stacking, uses simplified observation space.
\end{itemize}

Additionally, in order to compensate for any unfairness due to the reward shaping specific to Lucy-SKG, as well as to study further the impact of KRCs, we also trained Necto with Lucy-SKG's full reward design (namely \textbf{Necto-Reward}) and include these evaluation results as well.

In summary, in all cases it is evident that each component of Lucy-SKG makes it even more competent. The sole exception is when replacing the full observation space with the simplified space. However, this was to be expected; our intent was not to improve performance this way, but to reduce the amount of information received by the agent, plus reduce memory usage and computational complexity (as mentioned in Section~\ref{obsspace}). In particular, in our experiments we measured that using our simplified space would double the Frames Per Second (FPS) of the game during training (i.e. \textasciitilde4000 FPS) compared to using the full space (\textasciitilde2000 FPS) in our hardware. 5-action stacking however appears to balance the negative effect in the agent's learning performance caused by the simplified state space.

It is evident that the auxiliary tasks increase sample efficiency significantly. Moreover, KRCs appear to have a strong positive effect in all cases where they are present (Lucy or Necto-Reward) over linear combinations (e.g. Necto-Reward outperforms Necto). In addition, Lucy-SKG still manages to outperform the modified Necto model that used Lucy-SKG's reward methodology (i.e. Necto-reward), supporting further our claim that all Lucy-SKG components impact positively overall performance.

\section{Discussion}
\label{discussion}

Results indicate that RP provides substantial performance improvement, which is in line with our claim regarding immediate reward signals. However, SR is not performing equally well, which is consistent with the augmented baseline in \citet{jaderberg2016reinforcement}. Nonetheless, we believe that the reason for this is the design of the shared layers between auxiliary and main networks, which leaves most of the ``compressive" layers and by extent, the representations they hold inside the auxiliary network. Furthermore, we also argue that this simplistic shared architecture is generally very limited in terms of representation capacity. As observed in the results of Lucy---which uses a more complex attention-based shared architecture---auxiliary tasks increase the model's performance substantially.

It should be highlighted that due to constraints in computational resources we were unable to train models for more than 1B or 2B steps. In general, in-game behaviors of such models are not yet challenging for humans, which is why we did not evaluate Lucy-SKG against them. However, improving Lucy-SKG to outperform human experts is something we aspire to do in the future.

This work opens up new pathways in tackling challenging team-based control problems with complex dynamics and action space (e.g. aerial maneuvering), also incorporating sample-efficient techniques crucial for low computational-resource studies. In general, Lucy-SKG's components can be applied to any environment with teams (1 or more players each) competing within a bounded 2D or 3D geometrical space (a field) attempting to acquire/move an object (a ball) to a certain position (a goal).

\section{Conclusion}
\label{conclusion}

In this paper, we presented Lucy-SKG, a Deep Reinforcement Learning-based bot that learns to play Rocket League efficiently. We showed that with our various proposed novelties, Lucy-SKG managed to outperform Necto, the 2022 Rocket League Bot Champion, and its successor Nexto, with a significant difference. The array of novelties in parameterized reward shaping, and auxiliary representation learning enhanced model sample-efficiency and performance, which we showed empirically by evaluating our proposed methodology against appropriate baselines, as well as Necto and Nexto. Our discussion on these results also includes potential future extensions of our proposed methodology that could be applied in other problems as well.

\bibliographystyle{named}
\bibliography{neurips_2023}

\end{document}


\maketitle

\section{Environment Setup}
\label{environment}

In this section we provide details regarding the rewards used for Lucy-SKG, as well as for definition of the observation and action spaces.

\subsection{Rewards}\label{rewards_appendix}
\begin{figure}[t]
  \centering
  \includegraphics[width=1\linewidth]{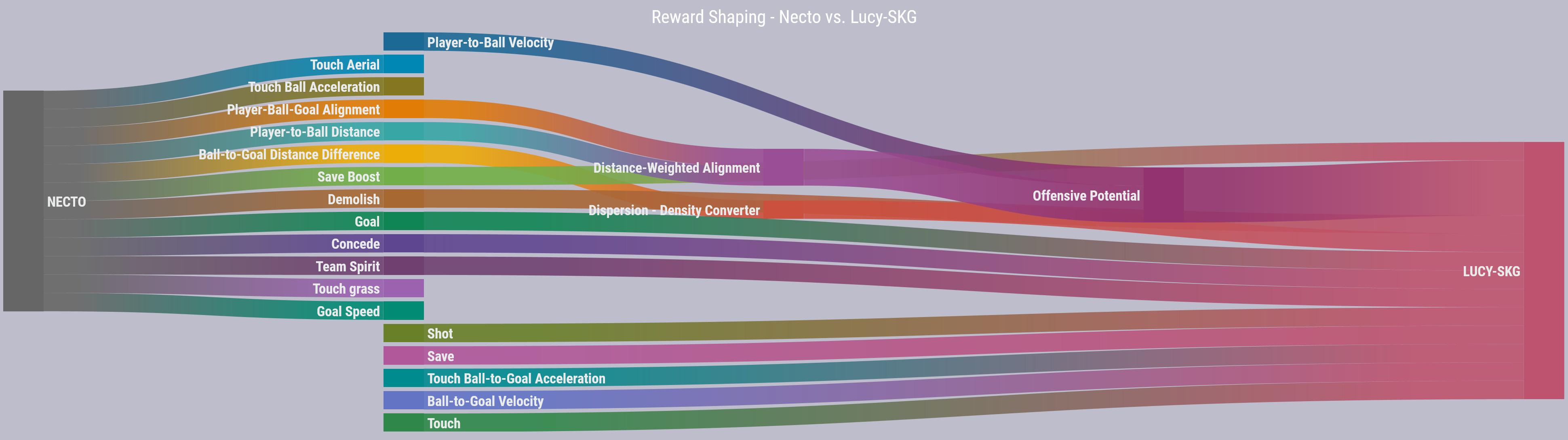}
  \caption{\centering Rewards used by Necto and Lucy-SKG.}
  \label{fig:rewards differences}
\end{figure}

RLGym reward functions draw inspiration from multiple works, such as the ones of \citet{liu2021motor} and \citet{berner2019dota}. Distance- and velocity-based rewards, in particular, are based on the work of \citet{liu2021motor}, but additionally utilize normalization constants in the form of maximum player and ball velocities. In Table \ref{tab:Lucy-SKG reward fn} we define all rewards used for Lucy-SKG, along with the rewards used for the auxiliary task learning ablation experiments. The rewards used by Lucy-SKG, as well as Necto's, are presented visually in Fig.~\ref{fig:rewards differences}.

The main characteristics of the reward function of Lucy-SKG can be described as follows:

\begin{itemize}
    \item \textbf{Use of KRCs.} Through KRCs, we introduce the:
    \begin{itemize}
        \item `Offensive Potential' reward function that indicates the offensive capability of the agent, by combining `Align Ball-to-Goal', `Player-to-Ball Distance' and `Player-to-Ball Velocity'.
        \item `Distance-weighted Alignment' reward function that indicates the quality of agent positioning, by combining `Align Ball-to-Goal' and `Player-to-Ball Distance'.
    \end{itemize}
    Parameterization of individual reward components was implemented for both `Offensive Potential' and `Distance-weighted Alignment'.
    \item \textbf{Distance reward parameterization.} `Ball-to-Goal Distance Difference' and `Distance-weighted alignment` employed were parameterized by dispersion and density, as defined in Eq. 3 in the main paper. For the former, a larger dispersion was given for positive reward distance compared to negative reward distance.
    \item \textbf{Careful selection of components.} All of the reward components used were carefully selected by analyzing game replays, visualizing rewards in the arena and computing desired minimum and maximum utility and event values.
    \item \textbf{Modification of reward functions.} `Touch Ball Acceleration' was replaced by `Touch Ball-to-Goal Acceleration' by introducing direction. The change penalizes hitting the ball toward the team goal and strongly rewards changing its direction towards the opponent goal.
\end{itemize}

\begin{figure}[t]
  \centering
  \includegraphics[width=0.8\columnwidth]{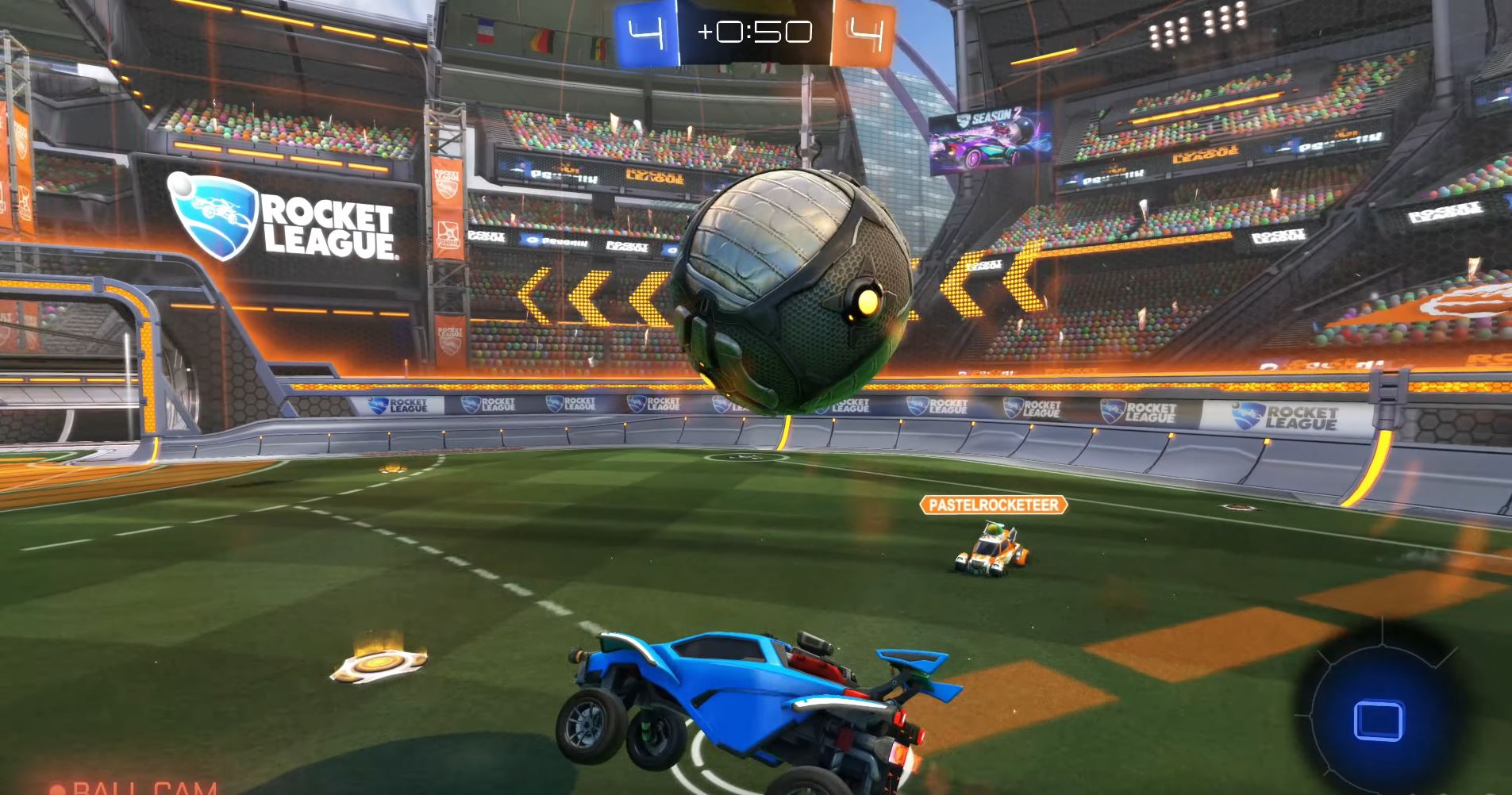}
  \caption{\centering Rocket League in-game screenshot.}
  \label{figure_1_rocket_league}
\end{figure}

A list of notations used for the reward definitions is given below. For simplicity and consistency reasons, we define the \textit{Blue} team to be tested agent's team (e.g. Lucy-SKG), and \textit{Orange} team to be the opponent's team (e.g. Necto).

\begin{itemize}
    \item $\ell_{goal}$: Goal depth
    \item $\teamGoalBackPosVec$: Back of own team goal (i.e. net) position
    \item $\goalBackPosVec$: Back of opponent team goal (i.e. net) position
    \item $\teamGoalCenterPosVec$: Own team goal center (i.e. goal line center)
    \item $\goalCenterPosVec$: Opponent team goal center (i.e. goal line center)
    \item $\ballPosVec$: Ball position
    \item $r_{ball}$: Ball radius
    \item $\overrightarrow{u}_{ball}$: Ball speed
    \item $\overrightarrow{\omega}_{ball}$: Ball angular velocity (radians)
    \item $\carPosVec$: Car position
    \item $\overrightarrow{u}_{car}$: Car speed
    \item $\objectDistance{i}{j} = \overrightarrow{p}_j - \overrightarrow{p}_i$: Euclidean distance between physics objects $i$ and $j$
    \item $w_{off}$: Offense weight for `Ball-to-Goal Distance Difference'
    \item $\offDispersionW$:  Offense dispersion for `Ball-to-Goal Distance Difference'
    \item $\offDensityW$: Offense density for `Ball-to-Goal Distance Difference'
    \item $w_{def}$: Defense weight for `Ball-to-Goal Distance Difference'
    \item $\defDispersionW$: Defense dispersion for `Ball-to-Goal Distance Difference'
    \item $\defDensityW$: Defense density for `Ball-to-Goal distance Difference'
    \item $\phi_{d_{p2b}}$: `Player-to-Ball Distance' KRC component
    \item $\phi_{u_{p2b}}$: `Player-to-Ball Velocity' KRC component
    \item $\phi_{a_{b2g}}$: `Align Ball-to-Goal' KRC component
    \item $boost$: Boost amount
    \item $\tau$: Team spirit factor for reward distribution
    \item $\mathcal{R'}_i$: Team spirit-distributed reward for player $i$
    \item $R'_i$: Shaped-MDP reward for player $i$
    \item $\bar{R'}_{team}$: Mean shaped-MDP team reward
    \item $\bar{R'}_{opponent}$: Mean shaped-MDP opponent reward
\end{itemize}

Additionally, parameter $\gamma$ in the reward shaping function $F$ was set to 1, due to utility reward differences being very small and turning otherwise negative.

Necto is an actively maintained project, constantly modified by its developers to further improve. However, although our research was initially flexible during the phase of study design, it was imperative to select a particular version of Necto for the later stages as a reference, as well as for performing consistent and concrete analysis and experiments. The latest version at the time, and which is the one we adopted, is the version committed in March 25, 2022\footnote{\url{https://github.com/Rolv-Arild/Necto/blob/2714729466551b9662b18898460cded6fdedb268/training/reward.py}}.

\begin{table*}[t]
    \centering
    \caption{Reward function components used for Lucy-SKG and auxiliary task learning ablations.}
    \label{tab:Lucy-SKG reward fn}
    \resizebox{\linewidth}{!}{
    \begin{tabular}{|c|c|c|c|c|}
    \hline
    \textbf{Type} & \textbf{Name} & \textbf{Weight} & \textbf{Formula}\\ \hline
    \multicolumn{4}{|c|}{\textbf{Lucy-SKG}}\\
    \hline
    \multirow{5}{*}{\begin{tabular}[c]{@{}c@{}}Reward\\ shaping\\ functions\end{tabular}} & \begin{tabular}[c]{@{}c@{}}Ball-to-Goal \\ Distance Difference\end{tabular} & \begin{tabular}[c]{@{}c@{}}offensive\\ dispersion: 0.6\\ defensive\\ dispersion: 0.4\\ weight: 2\end{tabular} &
    $
    \begin{aligned}
        \Phi_{d d_{b2g}} &= w_{off} * exp(-0.5 * \frac{\|\objectDistance{ball}{target}\| - \goalDepthVal}{6000 * \offDispersionW}^{1/\offDensityW} \\
        &\phantom{{} = } - w_{def} * \exp(-0.5 * \frac{\|\objectDistance{ball}{blue\ target}\| - \goalDepthVal}{6000 * \defDispersionW})^{1/\defDensityW}   
    \end{aligned}
    $
    \\ \cline{2-4} 
     & Ball-to-Goal Velocity & 0.8 & $\Phi_{u_{b2g}} = \frac{\objectDistance{ball}{target}}{\|\objectDistance{ball}{target}\|} \cdot \frac{\ballVelVec}{6000}$\\ \cline{2-4} 
     & Save boost & 0.5 & $\Phi_{boost} = \sqrt{boost/100}$\\ \cline{2-4} 
     & \begin{tabular}[c]{@{}c@{}}Distance-weighted \\Alignment \end{tabular} & \begin{tabular}[c]{@{}c@{}}dispersion: 1.1\\ weight: 0.6\end{tabular} & $\Phi_{dwa} = \|\phi_{a_{b2g}} * \phi_{d_{p2b}}\|^{1/2} * sgn(\phi)$ \\ \hline
     & Offensive Potential & \begin{tabular}[c]{@{}c@{}}density: 1.1,\\ weight: 1\end{tabular} & $\Phi_{op} = \|\phi_{a_{b2g}} * \phi_{d_{p2b}} * \phi_{u_{p2b}}\|^{1/3} * sgn(\phi)$\\ \hline
    \multirow{6}{*}{\begin{tabular}[c]{@{}c@{}}Event\\ reward\\ functions\end{tabular}} & Goal & 10 & $R_{goal} = \mathds{1}_{goal}$\\
     & Concede & -3 & $R_{concede} = \mathds{1}_{concede}$\\
     & Shot & 1.5 & $R_{shot} = \mathds{1}_{shot}$\\
     & \begin{tabular}[c]{@{}c@{}}Touch Ball-to-Goal\\ Acceleration\\\end{tabular} & 0.25 & $R_{touch_{ \mathrm{a}_{b2g}}} = \mathds{1}_{touch} * (r_{u_{b2g},t} - r_{u_{b2g}, t-1})$\\
     & Touch & 0.05 & $R_{touch} = \mathds{1}_{touch}$\\
     & Demolish & 2 & $R_{demo} = \mathds{1}_{demo}$\\
     & Demolished & -2 & $R_{demoed} = \mathds{1}_{demoed}$\\
    \hline
    \begin{tabular}[c]{@{}c@{}}Reward\\ distribution\end{tabular} & Team spirit & 0.3 & $\mathcal{R'}_i = (1 - \tau) * R'_i + \tau * \bar{R'}_{team} - \bar{R'}_{opponent}$\\
    \hline
    \multicolumn{4}{|c|}{\textbf{Auxiliary Task Learning Ablations}}\\
    \hline
    \multirow{5}{*}{\begin{tabular}[c]{@{}c@{}}Reward\\ functions\end{tabular}} & Player-to-Ball Velocity & 0.1 & $R_{u_{p2b}} = \frac{\objectDistance{car}{ball}}{\|\objectDistance{car}{ball}\|} \cdot \frac{\carVelVec}{2300}$\\
     & Ball-to-Goal Velocity & 1 & $R_{u_{b2g}} = \frac{\objectDistance{ball}{target}}{\|\objectDistance{ball}{target}\|} \cdot \frac{\ballVelVec}{6000}$\\
     & Team goal& 100 & $R_{team\ goal} = \mathds{1}_{team\ goal}$\\
     & Concede & 100 & $R_{concede} = \mathds{1}_{concede}$\\
     & Save & 30 & $R_{save} = \mathds{1}_{save}$\\
     & Shot & 30 & $R_{shot} = \mathds{1}_{shot}$\\
     & Demolish & 10 & $R_{demo} = \mathds{1}_{demo}$\\
    \hline
    \end{tabular}
    }
\end{table*}

\subsection{Observation Space}\label{observation_space_appendix}

A complete list of the observation features used by Lucy-SKG, compared to Necto's, is given in Table \ref{tab:Lucy-SKG_necto_features}.

Due to RLGym limitations in constructing the observation as a dictionary of three things, the observation triplet needed to be fit into a 2-d array, which the agent would subsequently decompose. The latent array was placed in the first row, while the byte array was placed in the rows below it. In addition, key padding mask booleans were represented by the last feature.

\begin{table}[h]
    \centering
    \caption{Comparison of observation features between Necto and Lucy-SKG along with \textit{Start} and \textit{End} positions on their respective observation vector. $k$ denotes the number of previous actions used for the action stacking ($k=5$ in our implementation).}
    \label{tab:Lucy-SKG_necto_features}
    \resizebox{0.6\textwidth}{!}{
    \begin{tabular}{|cc|c|cc|}
        \hline
        \multicolumn{2}{|c|}{\textbf{Necto}} & \textbf{Obs. Features} & \multicolumn{2}{c|}{\textbf{Lucy-SKG}}\\ \hline
        \multicolumn{1}{|c|}{\textbf{Start}} & \textbf{End} & \textbf{} & \multicolumn{1}{c|}{\textbf{Start}} & \textbf{End} \\ \hline
        \multicolumn{1}{|c|}{1} & 4 & \begin{tabular}[c]{@{}c@{}}main player, teammate,\\ opponent \& ball flags\end{tabular} & \multicolumn{1}{c|}{1} & 4 \\ \hline
        \multicolumn{1}{|c|}{5} & 5 & boost pad flag & \multicolumn{1}{c|}{-} & - \\ \hline
        \multicolumn{1}{|c|}{6} & 8 & \begin{tabular}[c]{@{}c@{}}normalized (relative)\\ linear position\end{tabular} & \multicolumn{1}{c|}{5} & 7 \\ \hline
        \multicolumn{1}{|c|}{9} & 11 & \begin{tabular}[c]{@{}c@{}}normalized (relative)\\ linear velocity\end{tabular} & \multicolumn{1}{c|}{8} & 10\\ \hline
        \multicolumn{1}{|c|}{12} & 14 & forward rotation vector & \multicolumn{1}{c|}{11} & 1\\ \hline
        \multicolumn{1}{|c|}{15} & 17 & upward rotation vector & \multicolumn{1}{c|}{14} & 16 \\ \hline
        \multicolumn{1}{|c|}{18} & 20 & angular velocity & \multicolumn{1}{c|}{17} & 19 \\ \hline
        \multicolumn{1}{|c|}{21} & 21 & normalized boost amount & \multicolumn{1}{c|}{20} & 20 \\ \hline
        \multicolumn{1}{|c|}{22} & 22 & demolition/boost timer & \multicolumn{1}{c|}{-} & - \\ \hline
        \multicolumn{1}{|c|}{23} & 24 & on ground, has flip & \multicolumn{1}{c|}{21}             & 22 \\ \hline
        \multicolumn{1}{|c|}{-} & - & demolished flag & \multicolumn{1}{c|}{23} & 23           \\ \hline
        \multicolumn{1}{|c|}{-} & - & previous action(s) (query only) & \multicolumn{1}{c|}{24} & 24 + 8$k$ \\ \hline
        \multicolumn{1}{|c|}{25} & 25 & key padding mask boolean & \multicolumn{1}{c|}{25 + 8$k$} & 25 + 8$k$ \\ \hline
    \end{tabular}
}
\end{table}

\subsection{Action Space}
\label{actionspace}

The action space for RLGym consists of 8 continuous or discrete actions, described in Table \ref{tab:action space}.

However, actions can be condensed or expanded to produce the original 8 through combination. An example of this is the keyboard-mouse action parser that both Necto and Lucy-SKG make use of. This means that both agents produce a binary/trinary 5-action keyboard and mouse output, which is transformed into the 8-action output that is required by RLGym. The same applies for Nexto, which, despite the fact it outputs 90 discrete actions, the actions map to an 8-action set.

\begin{table}[h]
    \centering
    \caption{The RLGym environment action space.}
    \label{tab:action space}
    \resizebox{0.6\textwidth}{!}{
    \begin{tabular}{|c|c|}
        \hline
        \textbf{Action} & \textbf{Properties}\\
        \hline
        \textit{Throttle} & \begin{tabular}[c]{@{}c@{}} -1 for full reverse, 1 for full forward.\\ Continuous or discrete.\end{tabular}\\
        \textit{Steer} & \begin{tabular}[c]{@{}c@{}}-1 for full left, 1 for full right\\. Continuous or discrete.\end{tabular}\\
        \textit{Pitch} & \begin{tabular}[c]{@{}c@{}}-1 for nose down, 1 for nose up.\\ Continuous or discrete.\end{tabular}\\
        \textit{Yaw} & \begin{tabular}[c]{@{}c@{}}-1 for full left, 1 for full right.\\ Continuous or discrete.\end{tabular}\\
        \textit{Roll} & \begin{tabular}[c]{@{}c@{}}-1 for full roll left, 1 for full roll right.\\ Continuous or discrete.\end{tabular}\\
        \textit{Jump} & \begin{tabular}[c]{@{}c@{}}0 for not jumping, 1 for jumping.\\ Discrete.\end{tabular}\\
        \textit{Boost} & \begin{tabular}[c]{@{}c@{}}0 for not using boost, 1 for using boost.\\ Discrete.\end{tabular}\\
        \textit{Handbrake} & \begin{tabular}[c]{@{}c@{}}0 for not using handbrake, 1 for using\\ handbrake. Discrete.\end{tabular}\\
        \hline
    \end{tabular}
    }
\end{table}

\section{Architectures}\label{architectures_appendix}

Our baseline architectures consist of the following:
\begin{enumerate}
    \item \textbf{Auxiliary task learning baseline}: To evaluate the efficacy of auxiliary task leaning methods, we used a simple Multi-Layer Perceptron (MLP) agent\footnote{\url{https://github.com/Impossibum/rlgym\_quickstart\_tutorial\_bot}}. This baseline agent consisted of 2 Fully Connected (FC) shared layers of 512 neurons each, intended for feature extraction, and 2 identical 3-layer 256-neuron branches for the actor and the critic.
    \item \textbf{Necto}: Necto uses two separate Perceiver-like architectures for the actor and the critic, with 2 2-layer 128-neuron MLPs for preprocessing the byte and latent arrays, and 2 Transformer encoder layers for performing 4-headed cross attention. Transformer encoder MLPs are 2-layer, with 512 hidden neurons and 128 output neurons. In between preprocessing and cross-attention, layer normalization is applied while the output is passed through ReLU.
    \item \textbf{Nexto}: Nexto's architecture is partly identical to Necto with the difference the action output is computed as the dot product between player and action embeddings. The player embedding is the regular Necto output passed through an additional linear layer, while the action embedding is computed through an 3-layer 32-neuron MLP for all 90 possible action outcomes. Thereby, the action that produces the highest score for the player is selected.
\end{enumerate}

\begin{figure*}[!t]
    \centering    \includegraphics[width=0.8\linewidth]
    {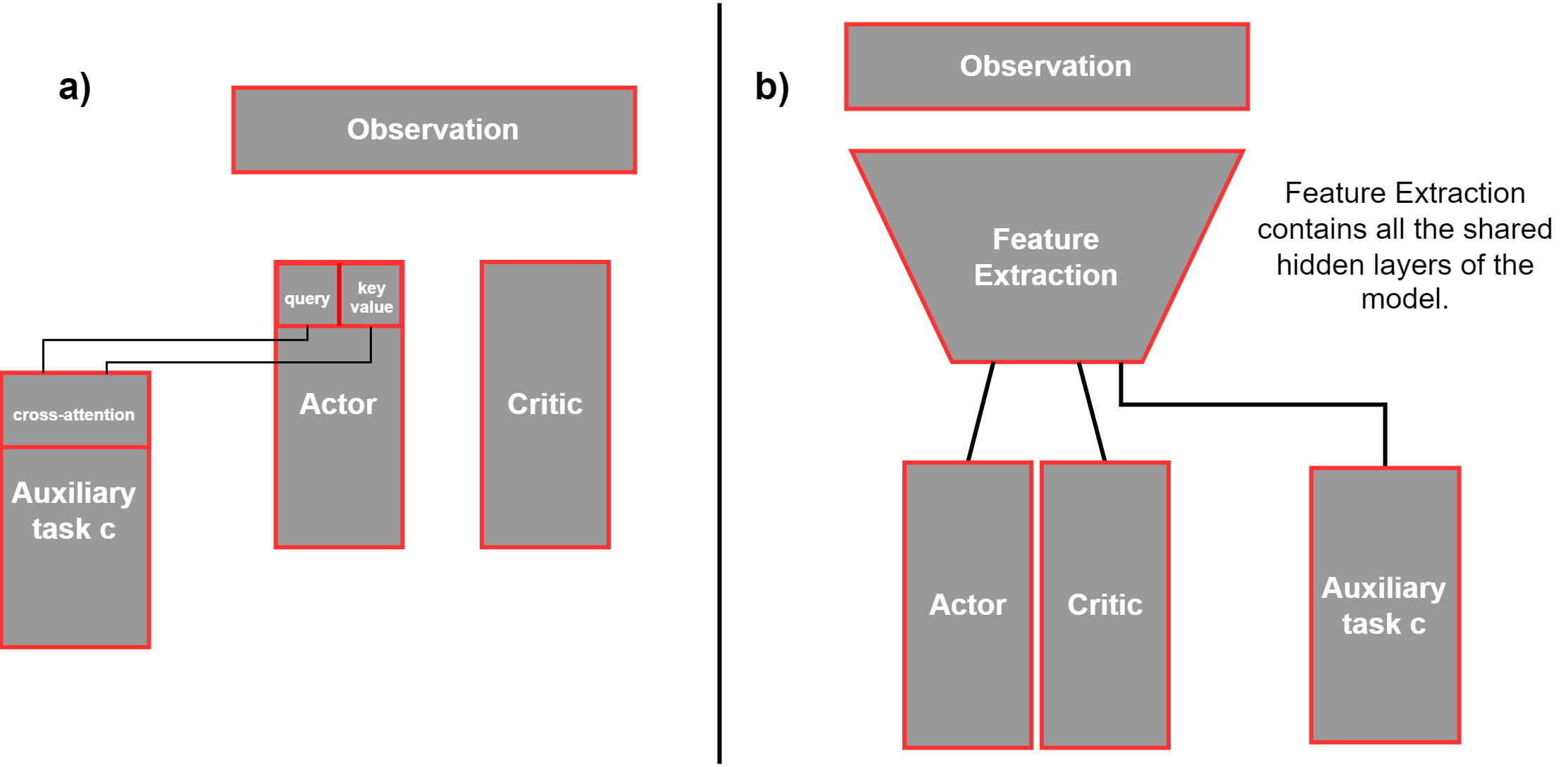}
	\caption{Auxiliary network is attached to a) Lucy-SKG architecture through the Actor network preprocessing layers b) the standalone auxiliary experiments architecture through the shared feature extraction layers of the model.
	}
	\label{fig:arch_design}
\end{figure*}

Our architecture designs are implemented as in Figure \ref{fig:arch_design} and include:
\begin{enumerate}
    \item \textbf{Reward Prediction (RP) Auxiliary Network}: Operates with a sequence length of 20. It has one LSTM layer with 32 recurrent units and a FC output layer. For auxiliary learning ablations, the RP network was connected to the baseline through branching from the shared layers. For Lucy-SKG, the network was connected to the actor only through the 2 preprocessing MLPs, the outputs of which were combined using a separate 4-headed cross-attention layer, specific to it. Following cross-attention, the single-query player dimension was squeezed and passed through the RP network. The RP network uses a categorical cross-entropy loss.
    \item \textbf{State Representation (SR) Auxiliary Network}: The encoder of the SR network has 3 FC layers with 128, 32 and 16 neurons respectively, with batch normalization and ReLU activation inbetween. The decoder begins as a mirrored version of the encoder, followed by a 512-neuron FC layer and another FC layer with neurons as many as the shape of a flattened observation. For connecting it to the auxiliary task learning baseline and Lucy-SKG, a procedure similar to the RP network was used. The SR network uses a smooth L1 reconstruction loss.
    \item \textbf{Lucy-SKG}: The main branch of the actor and the critic network of Lucy-SKG were based on the architecture of Necto, with the final ReLU layer missing.
\end{enumerate}


\section{RLBot \& RLGym}
\label{rlbot}
RLBot is an unofficial (yet endorsed by Psyonix) framework for creating Rocket League bots, allowing the growth of a community related to such projects. It supports bot development in various languages, including Python, which we used for our purposes. RLBot is mainly geared towards the development of hard-coded bots and some supervised learning models.

RLGym and RLBot are independent and have different mechanisms for accessing the game's internal state. Hence, we used RLGym for implementing and training Lucy-SKG as a Reinforcement Learning agent, and RLBot for evaluating their gameplay performance (Figure \ref{figure_1_rocket_league}).

\begin{figure}[t]
    \centering
    \begin{subfigure}[b]{0.49\columnwidth}
        \includegraphics[width=\linewidth]{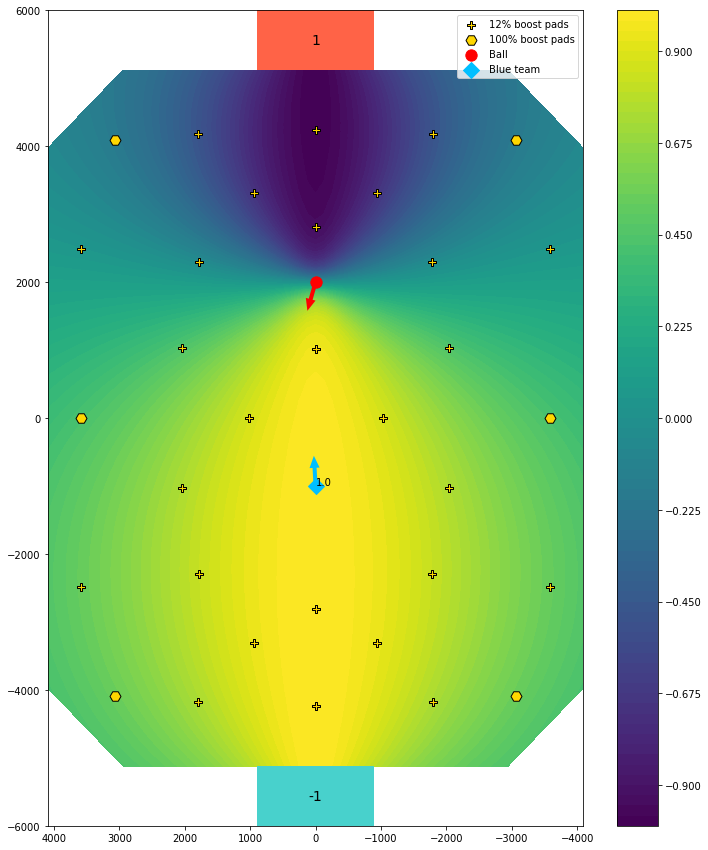}
    \end{subfigure}
    \begin{subfigure}[b]{0.5\columnwidth}
        \includegraphics[width=\linewidth]{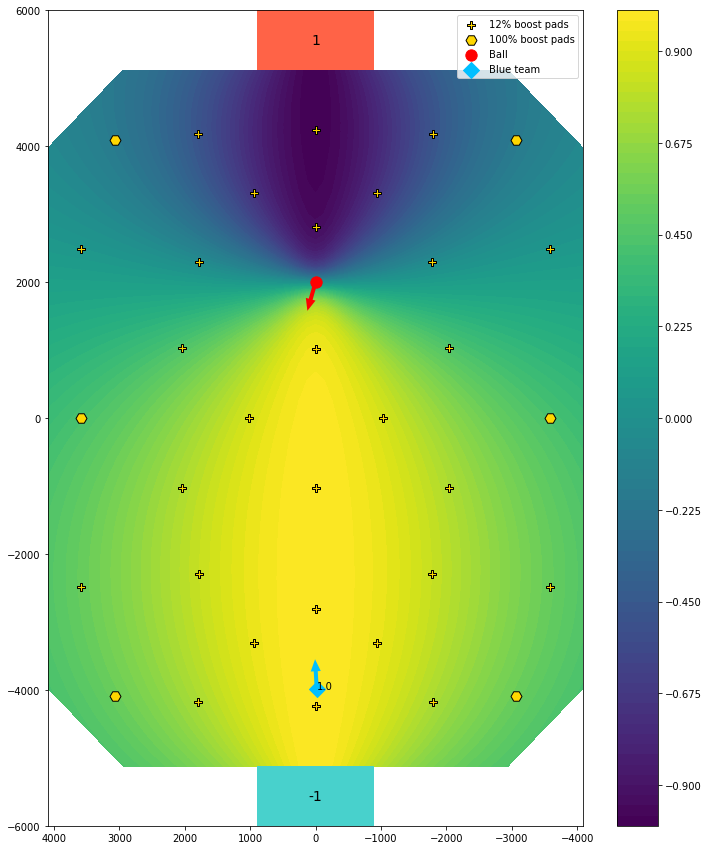}
    \end{subfigure}
    \caption{Existing `Align Ball-to-Goal' reward does not factor in player-to-ball distance, evident by the same beam-shaped reward distribution in two different cases.}
    \label{fig:align ball2goal}
\end{figure}

\section{Reward Analysis Library}
\label{rewardlibappendix}

\begin{figure*}[t]
    \centering
    \begin{subfigure}[b]{0.31\textwidth}
        \includegraphics[width=\textwidth]{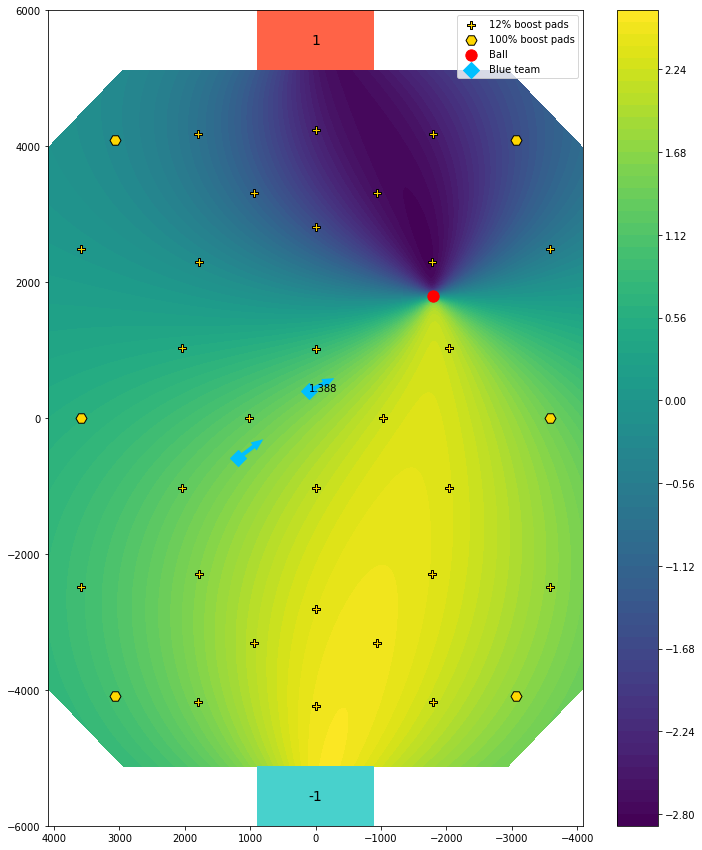}
        \label{fig:distributed reward}
    \end{subfigure}
    \begin{subfigure}[b]{0.31\textwidth}
        \includegraphics[width=\textwidth]{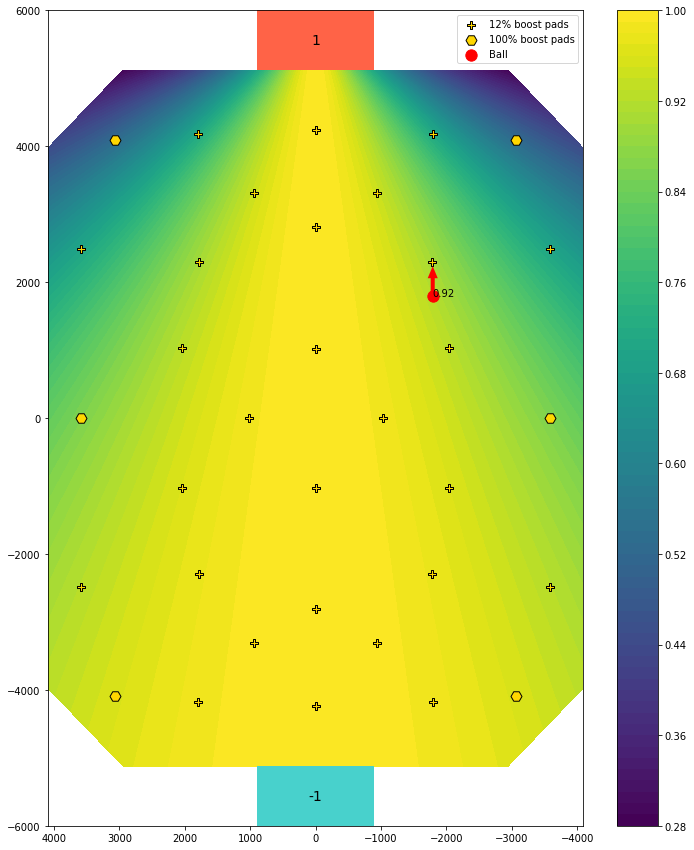}
        \label{fig:ball2goal vel}
    \end{subfigure}
        \begin{subfigure}[b]{0.31\textwidth}
        \includegraphics[width=\textwidth]{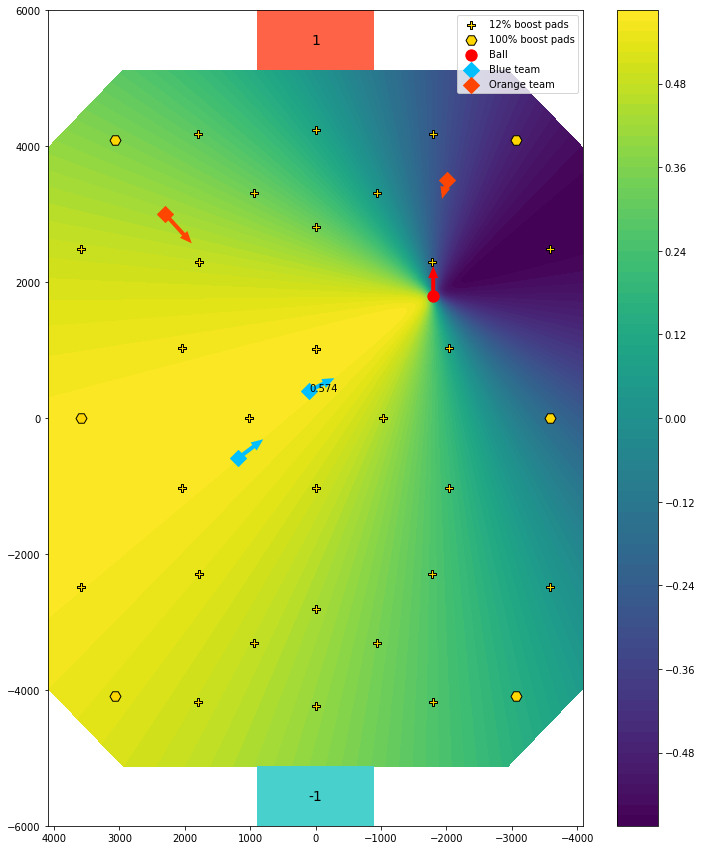}
        \label{fig:player2ball vel}
    \end{subfigure}
    \caption{Example reward plots generated by the rlgym-reward-analysis library. \textbf{Left:} Distributed reward combination of Align Ball-to-Goal and Player-to-Ball Distance rewards, with team spirit factor of $\tau=0.3$. \textbf{Center:} Ball-to-Goal Velocity reward. \textbf{Right:} Player-to-Ball Velocity reward.}
    \label{fig:example reward plots}
\end{figure*}

The reward analysis library was implemented in order to study existing RLGym reward functions (e.g. Figures \ref{fig:align ball2goal} and \ref{fig:example reward plots}), or custom reward functions, such as the ones presented in this work.
In this section we describe the technical details regarding the two modules available in the library, namely the \textit{Visualization module} and the \textit{replay file-to-reward} module.

\subsection{Visualization Module}
The first module, \plotArenaCode, allows developers to visualize reward functions provided to their models. The plot functionality is provided through the \plotArenaCode.\plottingCode.\arenaContourCode\ function, which plots a contour plot and receives the following parameters:
\begin{itemize}
    \item Basic parameters:
    \begin{itemize}
        \item \zCode: 1-d numpy array. Rewards values for each point in the arena. Refers to the \zCode\ parameter used by contour plots in Matplotlib.
        \item \ballPosCode: 3-d numpy vector, optional. Position of the ball in the arena.
        \item \ballVelCode: 3-d numpy vector, optional. Linear velocity of the ball.
        \item \playerPosCode: numpy array of shape $(n_{all}, 3)$ or 2-tuple of numpy arrays of shape $(n, 3)$. Optional. Player positions in the arena.
        \item \playerVelCode: Similar to \playerPosCode, optional. Player linear velocities.
    \end{itemize}
    \item Customization parameters:
    \begin{itemize}
        \item \goalWCode: \intCode\ or \floatCode, defaults to 1. Goal reward weight, used for annotation only.
        \item \playerIdxCode: \intCode\ or \noneCode, defaults to 0. The blue or orange team player index for which the rewards are plotted.\\
        If the player index is between 0 and $n_{blue} - 1$ a blue team player is annotated.\\
        If the player index is between $n_{blue}$ and $n_{blue} + n_{orange} - 1$ an orange player is annotated.\\
        If the player index is \noneCode\ no player is annotated.
        \item \annotateBallCode: \boolCode, defaults to \falseCode. Indicates whether the ball is annotated.
        \item \roundAnnotationCode: \intCode, defaults to 3. Number of floating point digits to round reward annotation to.
        \item \figsizeCode: \intCode\ or 2-tuple of \intCode s, defaults to $(12, 15)$. The size of the plot figure.
        \item \ballSizeCode: \intCode, defaults to 128. Ball marker size.
        \item \playerSizeCode: \intCode, defaults to 128. Player marker size.
        \item \boostPadSizeCode: \intCode, defaults to 80. Boost pad marker size.
        \item \contourLevelsCode: \intCode, defaults to 80. Number of contour plot regions.
    \end{itemize}
\end{itemize}

By importing \importCode\ \rlgymRewardAnalysisCode.\plotArenaCode.\plottingCode, the library initializes:
\begin{itemize}
    \item The arena using a Matplotlib Triangularization object, with a triangularization factor of 6.
    \item \arenaPositionsCode, with a fixed height set to 300. Arena positions are used for computing reward function values.
    \item A K-dimensional tree  containing all of the arena positions. The K-dimensional tree helps lookup points in the arena that are nearest to provided player positions and annotate players with the corresponding reward value.
\end{itemize}

Reward function values are computed as 1-d numpy arrays containing values for each position in the arena, through functions available through the \plotArenaCode.\rewardFnsCode\ module. Reward functions are divided into \commonRewardFnsCode, \extraRewardFnsCode\ --- reward functions provided by RLGym --- and \customRewardFnsCode --- novel reward functions introduced in this work.

Examples of plots generated by the rlgym-reward-analysis library can be seen in Figure A.3.

\subsection{Replay File-to-reward Parsing Module}

For the second module, the \parseReplayCode\ function accepts the following parameters:
\begin{itemize}
    \item \dfCode: Pandas Dataframe object. The dataframe of the replay for which to return reward values.
    \item \rewardNamesArgsCode: Sequence of reward name \stringCode s or 2-tuples of a reward name \stringCode\ and a dictionary of parameters.  Defaults to \noneCode.\\
    Available reward names can be found in \parseReplayCode.\rewardFnsCode.\rewardsNamesMap.\\
    If \rewardNamesArgsCode\ is \noneCode, \rewardNamesFnsCode\ is used instead.
    \item \rewardNamesFnsCode: Dictionary with reward name \stringCode\ keys and reward function callable values. If \rewardNamesArgsCode\ is \noneCode, \rewardNamesFnsCode\ should be provided.
\end{itemize}

The \parseReplaysCode\ function accepts the following parameters:
\begin{itemize}
    \item \folderPathsCode: Dictionary of replay group \stringCode\ keys and \stringCode\ sequences of folders values containing game replay CSVs.
    \item \rewardNamesArgsCode: Sequence of reward name \stringCode s or 2-tuples of reward name \stringCode s and dictionary of parameters.
    \item \nSkipCode: \intCode, defaults to 9. Number of replay file frames to skip in parsed replays.
\end{itemize}

\section{Experimental Setup}\label{experimental setup}

For our implementation, we used version 1.5.0 of Stable-Baselines 3 \citep{raffin2021stable}, a Reinforcement Learning library built on top of PyTorch as a backend, and RLGym version 1.1.0 \citep{rlgym}. In order for our reward function to work properly, this version of RLGym had to be modified, so as player and ball velocities would not become 0 when individual velocity-based reward components were computed.

For the reinforcement learning algorithm, we used a device-alternating variant of Proximal Policy Optimization (PPO) that alternates between transferring the model to the main memory, for gathering rollout buffer data, and transferring the model to the GPU memory, for training the agent. This implementation aided toward quicker experience collection, by eradicating data transfer between the CPU and the GPU, and reduced trainFor ing clock times, by utilizing the compute capacity of the GPU.

For comparison purposes, the state setter used was also similar to the one used in Necto\footnote{\url{https://github.com/Rolv-Arild/Necto/blob/1cf04ec5b67c5f6f5fc448d97a8e73ee2e15b630/training/state.py}}. The state setter serves so as to randomly reset the state of an instance after the episode has ended, with the following probabilities:
\begin{itemize}
    \item A real-world game-replay state of Platinum, Diamond, Champion, Grand Champion or Supersonic Legend rank with probability 0.7.
    \item A random state with probability 0.15.
    \item A kickoff state with probability 0.05.
    \item A kickoff-like state with probability 0.05.
    \item A goalie practice-like state, where one car is spawned near the goal for defense purposes, with probability 0.05.
\end{itemize}

The training environment varied between 10 to 20 game instances, handled by a separate process each, with 2v2 self-play. Additionally, for logging individual, episode-mean, unweighted reward components, we employed one ``logger'' match instance, out of a total of 10, as an indicative, validation-like environment. Rewards were logged for the blue team only, since many positive blue-team reward values can be negative for the orange team, and vice versa, bringing episode reward mean values close to 0.

Lastly, terminal conditions during training for each match were set to either 5 minutes of simulated gameplay, 45 seconds of no players touching the ball, or a goal being scored (Table \ref{tab:terminal conditions}).

\begin{table}[h]
    \centering
    \caption{Terminal conditions}
    {\renewcommand{\arraystretch}{1.2}
    \begin{tabular}{|c|}
        \hline
    \textbf{Terminal conditions}\\
        \hline
        5' of gameplay\\
        45'' of no players touching the ball\\ 
        Goal scored\\
        \hline
    \end{tabular}
    }
    \label{tab:terminal conditions}
\end{table}

\begin{table}[h]
    \centering
    \caption{
    Simulated gameplay years in terms of time steps. Years are computed using a frame skip of 8, i.e. 15 actions per simulated gameplay second.
    }
    {\renewcommand{\arraystretch}{1.2}
    \begin{tabular}{|c|c|}
        \hline
        \textbf{Years} & \textbf{Time steps}\\
        \hline
        1 & 473,040,000\\
        2 & 946,080,000\\
        3 & 1,419,120,000\\
        4 & 1,892,160,000\\
        5 & 2,365,200,000\\
        6 & 2,838,240,000\\
        7 & 3,311,280,000\\
        8 & 3,784,320,000\\
        9 & 4,257,360,000\\
        10 & 4,730,400,000\\
        \hline
    \end{tabular}
    }
    \label{tab:year timesteps}
\end{table}

For our training hyperparameters, we used the following:
\begin{itemize}
    \item 320,000 rollout steps, a batch size of 4,000 and a clip ratio of 0.2.
    \item The selected optimizer was Adam, with a learning rate of 0.0001.
    \item The entropy coefficient was set to 0, while the value function coefficient was set to 0.5.
    \item Regarding time granularity, all our experiments used a frame skip of 8. This means that, since the Rocket League physics engine runs at 120 Hz / second, or fps, 15 actions were performed, for 8 frames each, for every second of simulated gameplay (Table \ref{tab:year timesteps}).
    \item The discount value was set to approximately $\gamma \approx  0.995$, in order to achieve a $\gamma$ half-life --- $\gamma$ exponential reduced to 0.5 --- of 10 simulated gameplay seconds.
    \item For auxiliary losses, weights $\lambda_{SR}$ and $\lambda_{RP}$ were set to 1.
    \item For the RP auxiliary task, due to rewards never being 0 in practice, a threshold of 0.009 and -0.009, for positive and negative rewards respectively, was used to define zero-class rewards. The threshold was computed through rollout data analysis, with the goal of balancing positive-, negative- and zero-class rewards as much as possible.
\end{itemize}

\begin{table}[h]
    \centering
        \caption{
    Auxiliary task ablation hyperparameters.
    }
    {\renewcommand{\arraystretch}{1.2}
    \begin{tabular}{|c|c|}
        \hline
        \textbf{param} & \textbf{value}\\
        \hline
        epochs & 10\\
        learning rate & 5e-5\\
        entropy coef. & 0.01\\
        vf coef. & 1\\
        gamma & $\dfrac{e^{log(0.5)}}{fps * half\_life\_seconds}$\\
        batch\_size & 10\% of rollout\\
        n\_steps & 1M\\
        \hline
    \end{tabular}
    }
    \label{tab:aux_ppo_hyperparams}
\end{table}

\begin{table*}[!t]
  \centering
  \caption{Detailed numerical results from evaluation of Lucy-SKG trained for various steps, against Necto (1B and 2B steps). Percentage values denote the fraction of matches that Lucy-SKG was ahead in score.}
\resizebox{\textwidth}{!}{
\begin{tabular}{|c|c|c|c||c|c|}
\hline
\multicolumn{2}{|c|}{\textbf{Team 1 (Blue)}} & \multicolumn{4}{c|}{\textbf{Team 2 (Orange)}} \\
\hline
\textbf{Step (M)} &       & \textbf{Final Score (vs. Necto (1B))} & \multicolumn{1}{p{4.93em}||}{\textbf{Blue ahead in score \%}} & \textbf{Final Score (vs. Necto (2B))} & \multicolumn{1}{p{4.93em}|}{\textbf{Blue ahead in score \%}} \\
\hline
100   & \multirow{10}{*}{\textbf{Lucy-SKG}} & 8 - 300 & 0\%   & 5 - 300 & 0\% \\
\cline{1-1}\cline{3-6}200   &       & 294 - 300 & 36.93\% & 90 - 300 & 0.50\% \\
\cline{1-1}\cline{3-6}300   &       & 300 - 144 & 98.68\% & 110 - 300 & 0.45\% \\
\cline{1-1}\cline{3-6}400   &       & 300 - 234 & 26.14\% & 117 - 300 & 0\% \\
\cline{1-1}\cline{3-6}500   &       & 300 - 94 & 99.50\% & 300 - 258 & 100\% \\
\cline{1-1}\cline{3-6}600   &       & 300 - 68 & 100\% & 300 - 184 & 95.80\% \\
\cline{1-1}\cline{3-6}700   &       & 300 - 72 & 96.57\% & 300 - 217 & 75.19\% \\
\cline{1-1}\cline{3-6}800   &       & 300 - 59 & 100\% & 300 - 187 & 67.27\% \\
\cline{1-1}\cline{3-6}900   &       & 300 - 71 & 100\% & 300 - 171 & 94.44\% \\
\cline{1-1}\cline{3-6}1000  &       & 300 - 54 & 100\% & 300 - 134 & 99.09\% \\
\hline
\end{tabular}%
}
  \label{tab:evaluation results detailed}%
\end{table*}%

Other design choices present in our work are elaborated as follows:

\begin{itemize}
\item
  \textbf{State representation}: We reconstruct the key/value objects only of Lucy-SKG's observation because it represents all of the observation minus previous player actions with padding not being a
  part of the observation and the player query being only a part of it.
\item
  \textbf{Auxiliary tasks in actor only}: Auxiliary tasks were employed for the actor only to reduce backpropagation cost with the goal of creating an equally effective and efficient agent.
\item
  \textbf{Hyperparameter choices}: Most of the parameters were selected manually through trial and error, due to the lack of computational resources to search for better values. Batch size could not be
  increased further again due to lack of memory (our batch of size 4000 requires \textasciitilde16gb per batch sequence). Gamma was computed for a half-life (exponent reduced to 0.5) of 10 seconds so as the agent can be greedy enough to make goals as early as possible but also be farsighted.
\end{itemize}

\subsection{Auxiliary Task Ablations Setup}
The training environment consisted of 10 game instances in total, with each instance handled by a separate process. Matches were set to 1v1 with self-play, $\gamma$ was computed in order to achieve a half-life of 5 simulated seconds, and terminal episode conditions were either a scored goal or 1000 steps (approximately $\sim$66 seconds, using 
8 frame skips). The experiments ran for 500M steps each. Hyper-parameters of PPO were set as shown in Table \ref{tab:aux_ppo_hyperparams}.

Moreover, for RP task ablations, the zero-reward threshold was set to 0.005 and -0.005, for positive and negative rewards respectively.

\subsection{Hardware Equipment}

Our experiments were performed on two machines:

\begin{enumerate}
    \item A primary machine with an i9-12900K CPU, 2 Nvidia RTX 3090 GPUs and 64 GB RAM. Game graphics would run on the second GPU, while the first one was used to train the agent.
    \item A secondary machine with an i7-8700 CPU, an Nvidia Titan V GPU and 32 GB RAM. This machine was used for running initial experiments and to evaluate the efficacy of auxiliary task learning methods.
\end{enumerate}

\section{Complementary Experimental Results}
\label{complementary exp res}

In this section we provide further details regarding training and evaluation results.

\subsection{Evaluation Results}

In Table \ref{tab:evaluation results detailed} we provide detailed numerical results regarding 20 sets of 300 independent one-goal head-to-head evaluation games we performed between Lucy-SKG trained for various steps, versus Necto trained for 1 billion and 2 billion steps. 

\subsection{Training Results}

\begin{figure*}
    \centering
    \begin{subfigure}[b]{0.31\textwidth}
        \includegraphics[width=\textwidth]{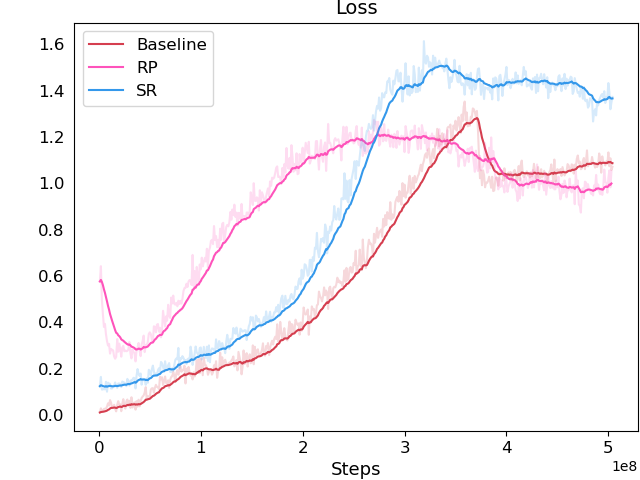}
        \label{fig:aux loss}
    \end{subfigure}
    \begin{subfigure}[b]{0.31\textwidth}
        \includegraphics[width=\textwidth]{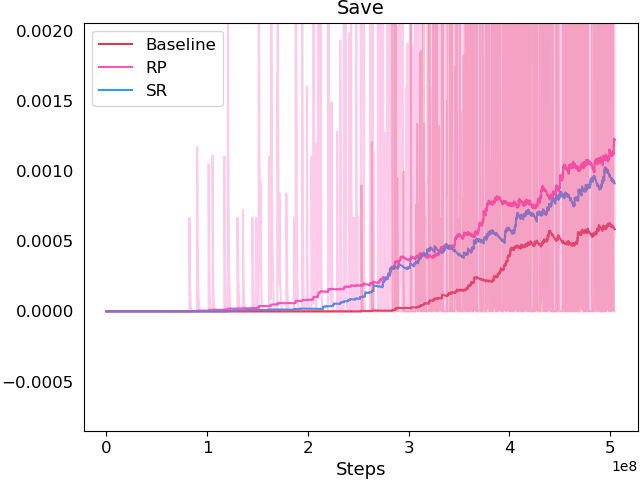}
        \label{fig:aux save}
    \end{subfigure}
        \begin{subfigure}[b]{0.31\textwidth}
        \includegraphics[width=\textwidth]{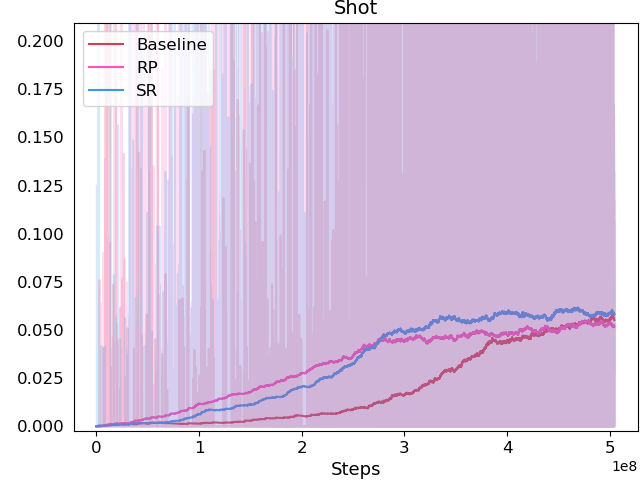}
        \label{fig:aux shot}
    \end{subfigure}
    \begin{subfigure}[b]{0.31\textwidth}
        \includegraphics[width=\textwidth]{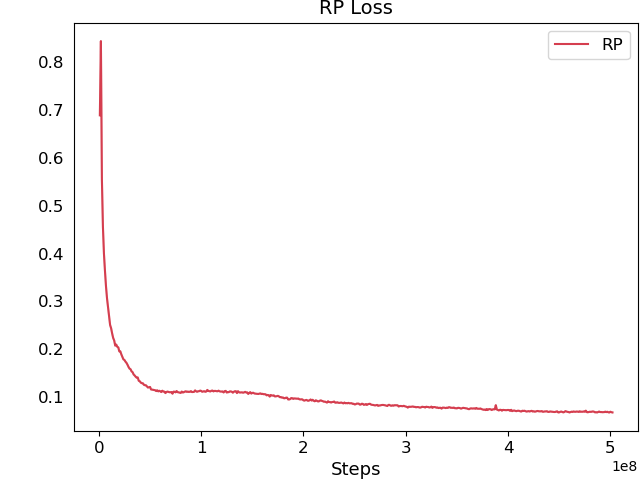}
        \label{fig:aux rp loss}
    \end{subfigure}
    \begin{subfigure}[b]{0.31\textwidth}
        \includegraphics[width=\textwidth]{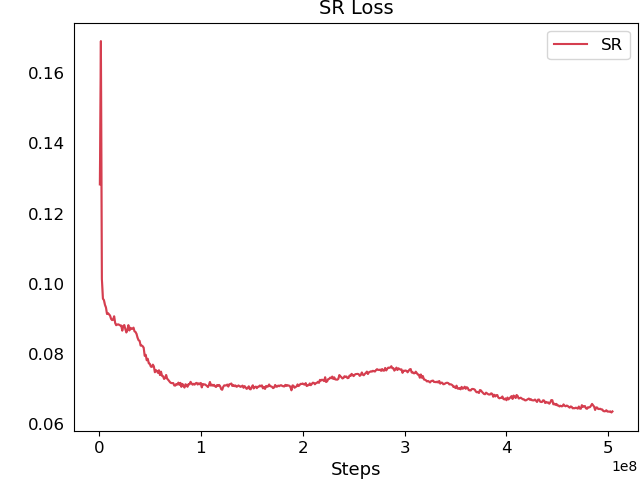}
        \label{fig:aux sr loss}
    \end{subfigure}
    \begin{subfigure}[b]{0.31\textwidth}
        \includegraphics[width=\textwidth]{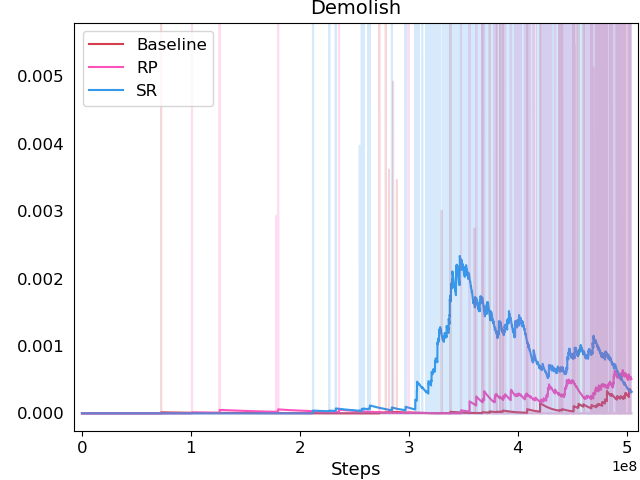}
        \label{fig:aux demolish}
    \end{subfigure}
        \begin{subfigure}[b]{0.31\textwidth}
        \includegraphics[width=\textwidth]{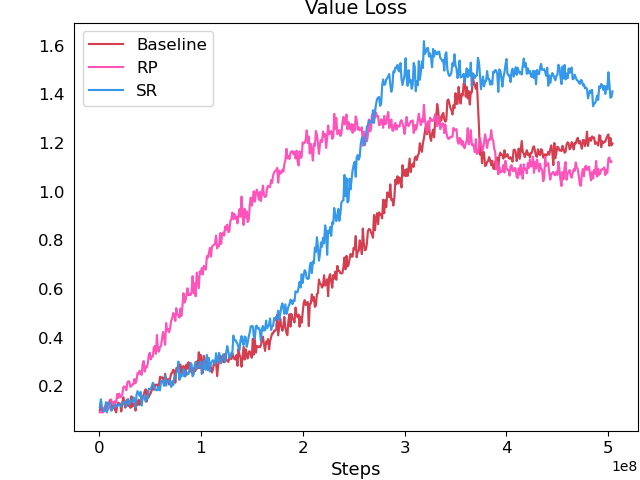}
        \label{fig:aux value loss}
    \end{subfigure}
    \begin{subfigure}[b]{0.31\textwidth}
        \includegraphics[width=\textwidth]{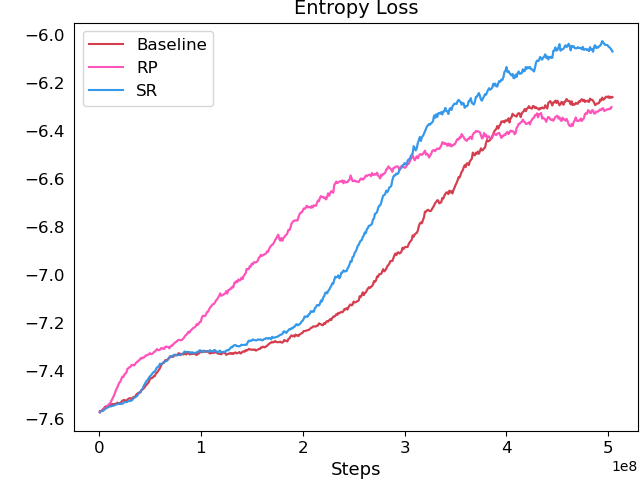}
        \label{fig:aux entropy loss}
    \end{subfigure}
        \begin{subfigure}[b]{0.31\textwidth}
        \includegraphics[width=\textwidth]{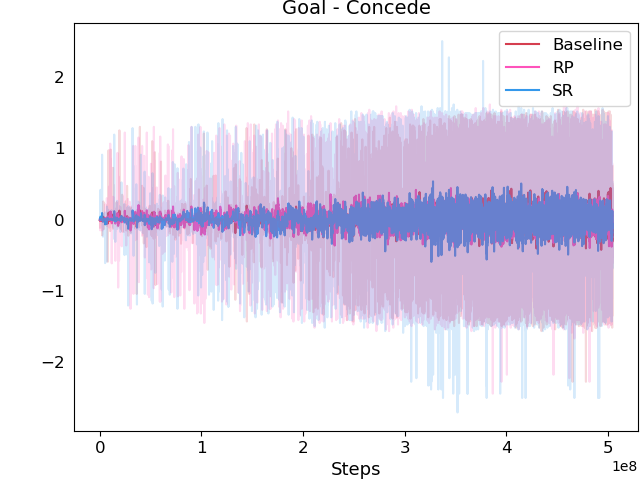}
        \label{fig:aux goal concede}
    \end{subfigure}
        \begin{subfigure}[b]{0.31\textwidth}
        \includegraphics[width=\textwidth]{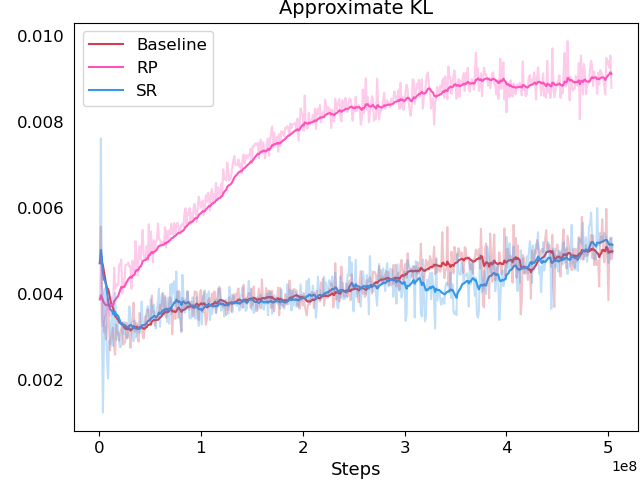}
        \label{fig:aux kl}
    \end{subfigure}
    \begin{subfigure}[b]{0.31\textwidth}
        \includegraphics[width=\textwidth]{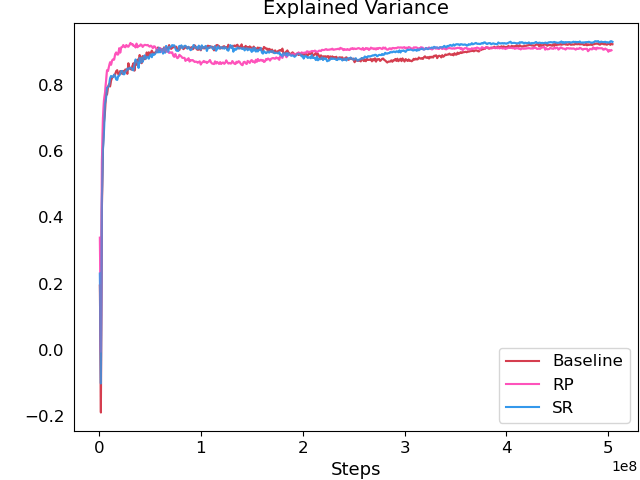}
        \label{fig:aux explained variance}
    \end{subfigure}
        \begin{subfigure}[b]{0.31\textwidth}
        \includegraphics[width=\textwidth]{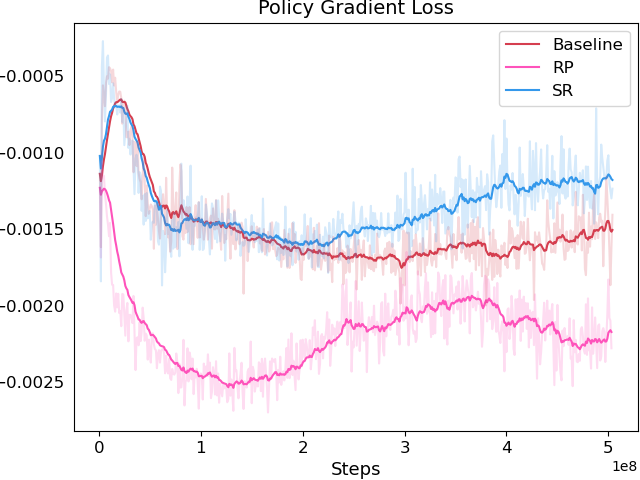}
        \label{fig:aux pg loss}
    \end{subfigure}
    \caption{Metrics and episode-mean rewards during training of auxiliary task models (SR and RP tasks) against the baseline. Exponential smoothing was used to highlight the learning trends during training. Shaded areas represent true (non-smoothed values).}
    \label{fig:aux figures}
\end{figure*}

\begin{figure*}
    \centering
    \begin{subfigure}[b]{0.31\textwidth}
        \includegraphics[width=\textwidth]{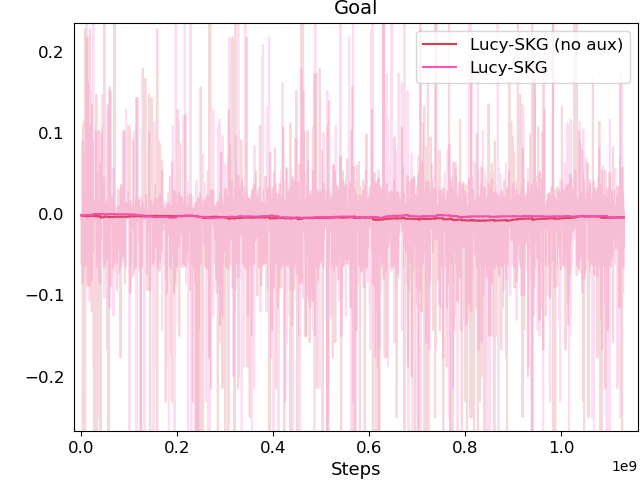}
        \label{fig:lucy goal}
    \end{subfigure}
    \begin{subfigure}[b]{0.31\textwidth}
        \includegraphics[width=\textwidth]{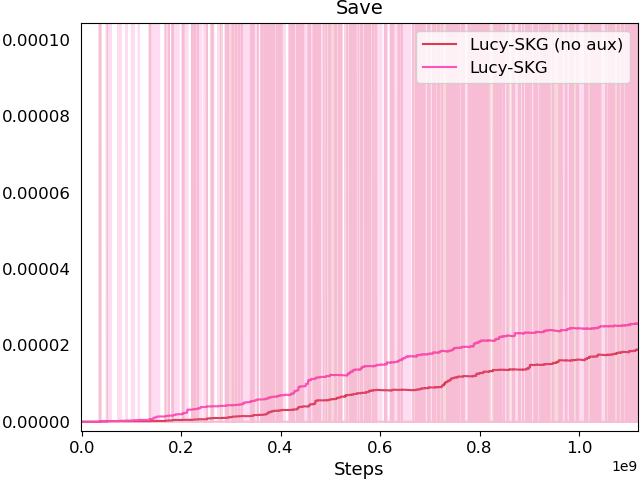}
        \label{fig:lucy save}
    \end{subfigure}
        \begin{subfigure}[b]{0.31\textwidth}
        \includegraphics[width=\textwidth]{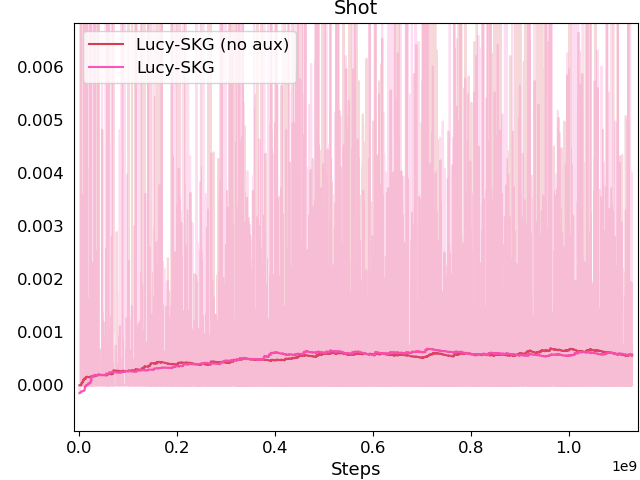}
        \label{fig:lucy shot}
    \end{subfigure}
    \begin{subfigure}[b]{0.31\textwidth}
        \includegraphics[width=\textwidth]{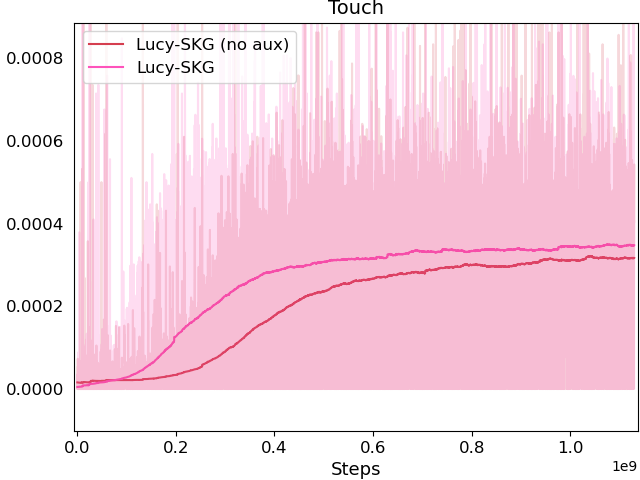}
        \label{fig:lucy touch}
    \end{subfigure}
    \begin{subfigure}[b]{0.31\textwidth}
        \includegraphics[width=\textwidth]{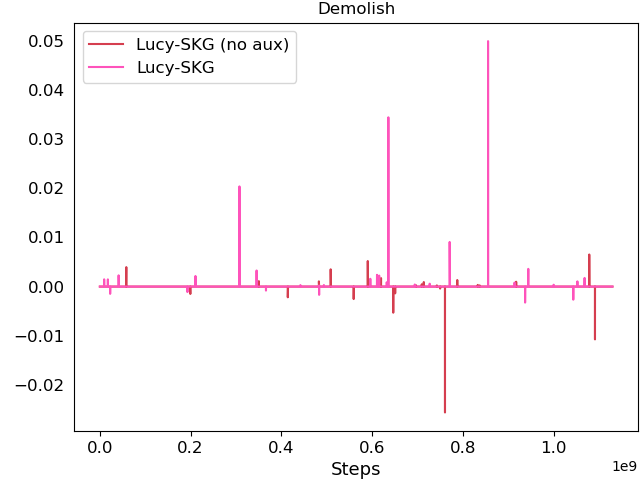}
        \label{fig:lucy demolish}
    \end{subfigure}
    \begin{subfigure}[b]{0.31\textwidth}
        \includegraphics[width=\textwidth]{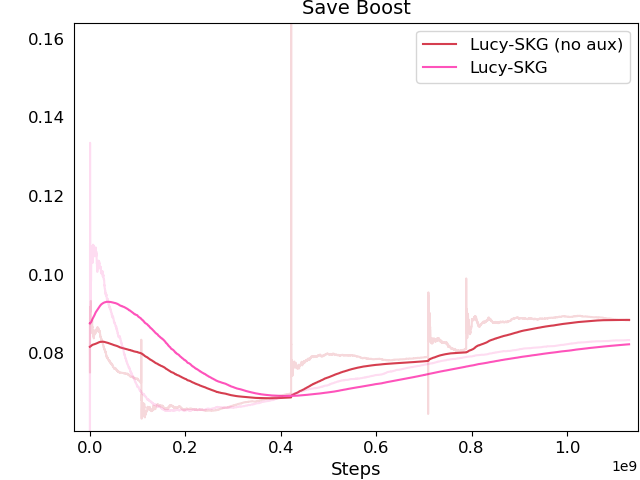}
        \label{fig:save boost}
    \end{subfigure}
        \begin{subfigure}[b]{0.31\textwidth}
        \includegraphics[width=\textwidth]{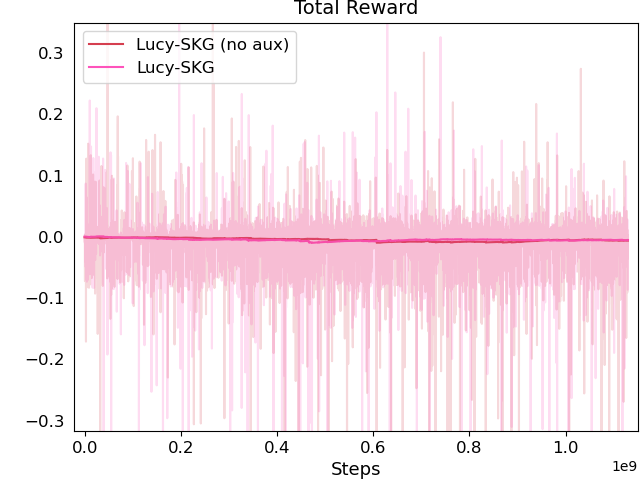}
        \label{fig:total reward}
    \end{subfigure}
    \begin{subfigure}[b]{0.31\textwidth}
        \includegraphics[width=\textwidth]{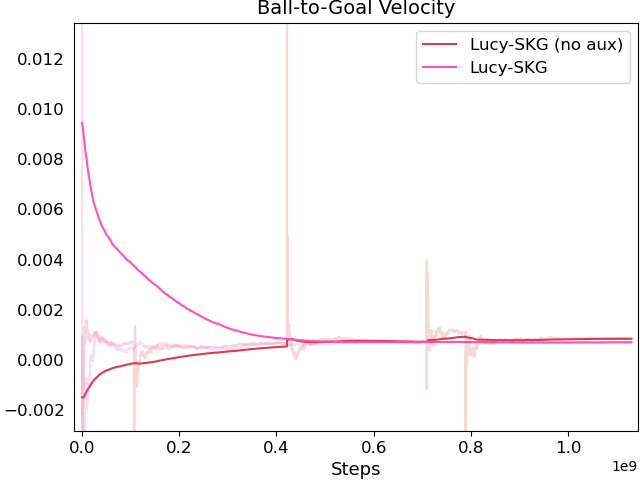}
        \label{fig:lucy b2g v}
    \end{subfigure}
        \begin{subfigure}[b]{0.31\textwidth}
        \includegraphics[width=\textwidth]{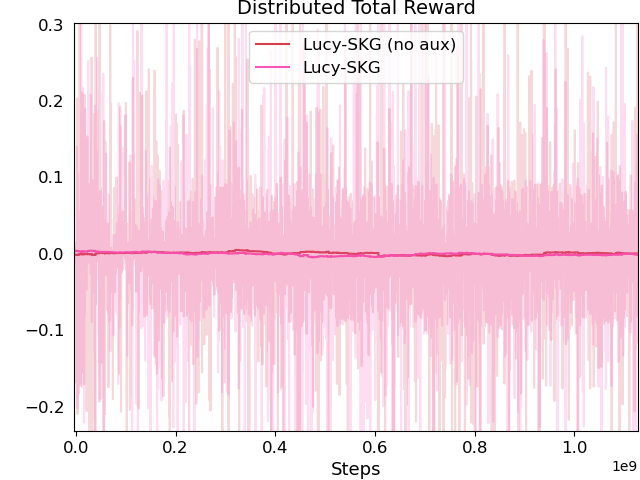}
        \label{fig:distributed total reward}
    \end{subfigure}
        \begin{subfigure}[b]{0.31\textwidth}
        \includegraphics[width=\textwidth]{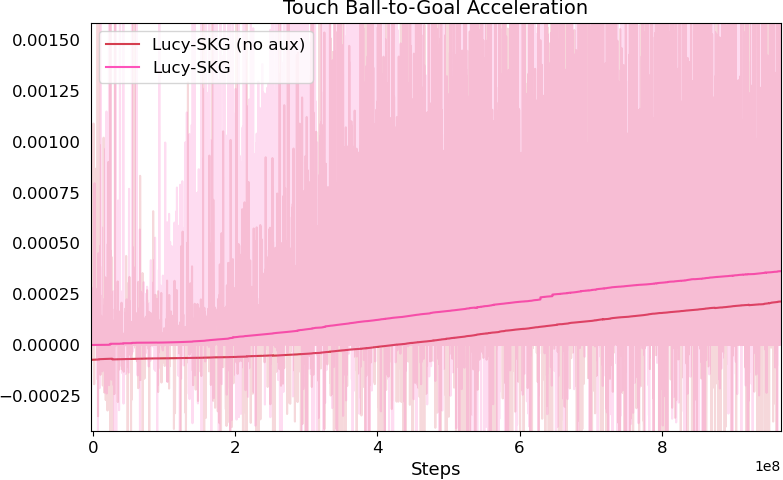}
        \label{fig:touch b2g accelera}
    \end{subfigure}
    \begin{subfigure}[b]{0.31\textwidth}
        \includegraphics[width=\textwidth]{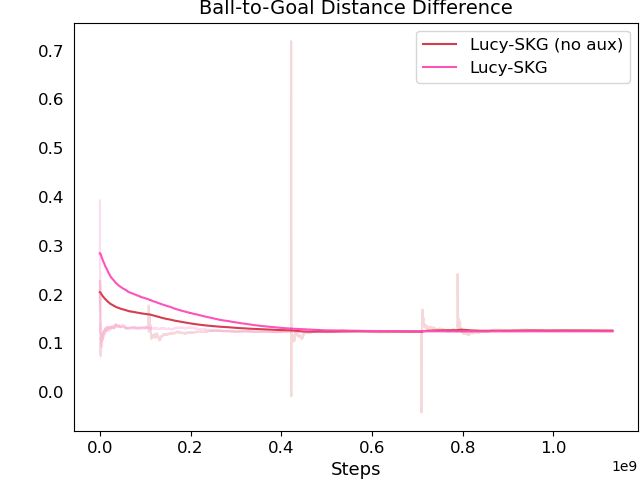}
        \label{fig:b2g dist diff}
    \end{subfigure}
        \begin{subfigure}[b]{0.31\textwidth}
        \includegraphics[width=\textwidth]{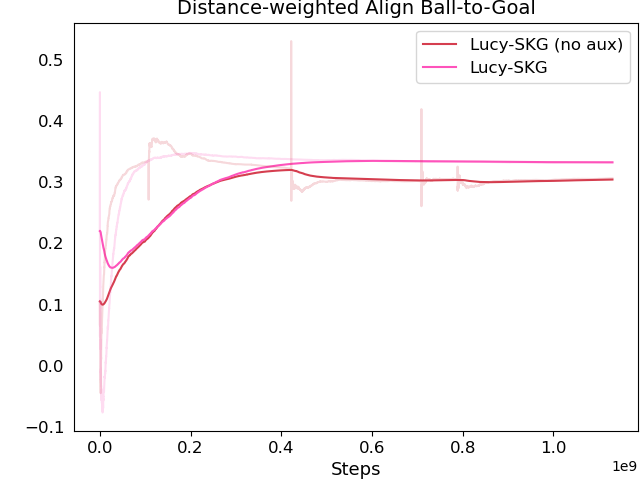}
        \label{fig:dist weight align b2g}
    \end{subfigure}
    \caption{Episode-mean rewards during training of Lucy-SKG and Lucy-SKG (no aux). Exponential smoothing was used to highlight the learning trends during training. Shaded areas represent true (non-smoothed values).}
    \label{fig:lucy skg train figures}
\end{figure*}

In this section we present results regarding rewards and metrics during training for the auxiliary task learning methods alone (Figure \ref{fig:aux figures}), and for Lucy-SKG with and without the use of auxiliary task learning methods (Figure \ref{fig:lucy skg train figures}). Although high variance is evident, this is expected due to the high complexity of the environment, giving rise to a lot of uncertainty (and at the same time, room for improvement) at these stages of training. Positive learning trends are evident in several cases, even if the scale of improvement is small.

\section{Graph representation}

In this section, we provide a proof-of-concept for treating the observation of the game as a graph. It has not yet been implemented for Lucy-SKG, but is left as future work and is described here so as to provide further insights on our methodology's potential extensions and use cases. 
A graph observation space allows for Graph Neural Networks (GNNs) to be employed as part of the processing by convolving nearby or similar objects for additional spatial information.

The proposed solution to this is the employment of an `Object-to-Object Distance` reward function, similar to `Player-to-Ball Distance` offered by RLGym. Furthermore, the reward function is parameterized by dispersion and density (Section 4.4), creating a parameterizable graph observation.

We present two cases regarding self-connections: a) a self-connection of weight 1 and b) a normalized self-connection. Both cases create non-symmetric adjacency matrices with certain advantages and disadvantages.

\paragraph{Case a) - self-connection of weight 1:}
\begin{equation}
    \label{eqn:adj matrix self-connection 1}
    \begin{gathered}
        \adjMatrix_{i,j} \leftarrow
        \begin{cases}
            \exp(-0.5 * \frac{\|\objectDistance{i}{j}\|}{2300 * \dispersionW})^{1/\densityW},\; i \neq j\\
            1,\; i = j
        \end{cases}\\
        \bar{\adjMatrix_i} \leftarrow
        \begin{cases}
            \frac{\sum_j \adjMatrix_{i,j}}{N},\; i \neq j\\
            1,\; i = j
        \end{cases}\\
        \adjMatrix \leftarrow \adjMatrix / \bar{\adjMatrix},
    \end{gathered}
\end{equation}

\paragraph{Case b) - normalized self-connection:}
\begin{equation}
    \label{eqn:adj matrix regular self-connection}
    \begin{gathered}
        \adjMatrix_{i,j} \leftarrow \exp(-0.5 * \frac{\|\objectDistance{i}{j}\|}{2300 * \dispersionW})^{1/\densityW}\\
        \bar{\adjMatrix_{i,j}} = \frac{\sum_j \adjMatrix_{i,j}}{N}\\
        \adjMatrix \leftarrow \adjMatrix / \bar{\adjMatrix}
    \end{gathered}
\end{equation}

where $i$ and $j$ are objects, $N$ is the total number of objects, $\adjMatrix$ is the adjacency matrix and $\bar{\adjMatrix_i}$ is the normalization matrix.

In case a), a self-connection weight of 1 means the self is always treated the same way and that certain objects that are nearby may become more important. In case b), a normalized self-connection means that the self will always be more important compared to objects nearby since it has a distance of 0. When other objects are too far, the self is attributed a disproportionately large weight.


\bibliographystyle{named}
\bibliography{neurips_2023}